\documentclass{article}
\usepackage{amsmath}
\usepackage{amssymb}
\usepackage{graphicx}
\usepackage{authblk}
\usepackage{tabu}
\usepackage[utf8]{inputenc}
\usepackage[]{algorithm2e}
\usepackage[colorlinks = true,
linkcolor = blue,
urlcolor  = blue,
citecolor = blue,
anchorcolor = blue]{hyperref}
\newcommand{\PX}{\mathcal{P}_\mathcal{X}}
\newcommand{\PY}{\mathcal{P}_\mathcal{Y}}
\newcommand{\x}{\mathbf x}
\newcommand{\y}{\mathbf y}
\newcommand{\R}{\mathbb{R}}
\newcommand{\E}{\mathbb{E}}
\newcommand{\RR}{\mathbf{R}}
\usepackage{amsthm}
\providecommand{\keywords}[1]{\textbf{Keywords:} #1}
\theoremstyle{plain}

\theoremstyle{definition}

\newtheorem{proposition}{Proposition}

\usepackage{geometry}
\usepackage[dvipsnames,svgnames]{xcolor}
\usepackage[normalem]{ulem}

\title{Targeting Solutions in Bayesian Multi-Objective Optimization: Sequential and Batch Versions\footnotetext{This is the author's version of the work. It is posted here for your personal use. Not for redistribution. The definitive Version of Record was published in Annals of Mathematics and Artificial Intelligence, volume 88, pp. 187-212 (2020), \url{https://doi.org/10.1007/s10472-019-09644-8.}}}
\author[1,2]{David Gaudrie}
\author[2,3]{Rodolphe le Riche}
\author[4]{Victor Picheny}
\author[1]{Benoît Enaux}
\author[1]{Vincent Herbert}
\affil[1]{Groupe PSA}
\affil[2]{École Nationale Supérieure des Mines de Saint-Étienne}
\affil[3]{CNRS LIMOS}
\affil[4]{INRA Toulouse}
\date{}

\begin{document}
	
	\maketitle
	
	\begin{abstract}
		Multi-objective optimization aims at finding trade-off solutions to conflicting objectives. These constitute the Pareto optimal set. In the context of expensive-to-evaluate functions, it is impossible and often non-informative to look for the entire set. 
		As an end-user would typically prefer a certain part of the objective space, we modify the Bayesian multi-objective optimization algorithm which uses Gaussian Processes and works by maximizing the Expected Hypervolume Improvement, to focus the search in the preferred region.
		The cumulated effects of the Gaussian Processes and the targeting strategy lead to a particularly efficient convergence to the desired part of the Pareto set. To take advantage of parallel computing, a multi-point extension of the targeting criterion is proposed and analyzed.
	\end{abstract}
	
	\keywords{Gaussian Processes \and Bayesian Optimization \and Computer Experiments \and Preference-Based Optimization \and Parallel Optimization}
	
	\section{Introduction}
	Multi-objective optimization (MOO) aims at minimizing $m$ objectives simultaneously: $\underset{\x\in X\subset\R^d}{\min}(f_1(\x),\dotsc,f_m(\x))$. As these objectives are generally competing, the optimal trade-off solutions known as the Pareto optimal set $\PX$ are sought. 
	These solutions are \emph{non-dominated}: it is not possible to improve one objective without worsening another; $\forall \x^* \in \PX$, $ \nexists \mathbf z \in X$ such that $\mathbf f(\mathbf z)\prec\mathbf f(\mathbf x^*)$ where ``$\prec$'' stands for Pareto domination.
	The image of $\PX$ in the objective space is called the Pareto front, $\PY=\{\mathbf f(\x),\x\in\PX\}$.
	The Ideal and the Nadir points bound the Pareto front and are defined respectively as $\mathbf I=(\underset{\y\in\PY}{\min}y_1,\dotsc,\underset{\y\in\PY}{\min}y_m)$ and $\mathbf N=(\underset{\y\in\PY}{\max}y_1,\dotsc,\underset{\y\in\PY}{\max}y_m)$. More theory and concepts in multi-objective optimization can be found in \cite{sawaragi1985theory,miettinen1998nonlinear}.
	
	MOO algorithms aim at constructing the best approximation to $\PY$, called the \emph{empirical }Pareto front (or approximation front) $\widehat{\PY}$ which is made of non-dominated observations. 
	The construction is of course iterative. The contribution of the sampled points is measured by different criteria, the ones based on the dominated hypervolume being state-of-the-art when searching for the entire Pareto front \cite{TheseZitzler,SMS,beume2007sms}.
	At the end of the search, $\widehat{\PY}$ is delivered to a Decision Maker (DM) who will choose a solution.
	
	However, when dealing with expensive computer codes, only a few designs $\x$ can be evaluated. In Bayesian optimization, a surrogate for each objective, $Y_j(\cdot)$, is first fitted to an initial Design of Experiments (DoE) evaluated at $n$ locations, $\mathcal{D}_j^n:=\{(\x^{(1)},f_j(\x^{(1)})),\dotsc,(\x^{(n)},f_j(\x^{(n)}))\}$, using Gaussian Processes (GP) \cite{EGO}. 
	Classically, to contain the computational complexity, the metamodels $Y_j(\cdot)$ are assumed to be independent. In \cite{TheseSvensson} dependent GPs have been considered without noticing significant benefits.
	Information given by $\mathbf Y(\cdot):=(Y_1(\cdot),\dotsc,Y_m(\cdot))^\top$ is used to sequentially evaluate new promising inputs with the aim of reaching the Pareto front \cite{Parego,Keane,EHI,SMS,EMI,SUR}. 
	It is now established, e.g. in \cite{SUR}, that the algorithms with embedded metamodels such as GPs call much more sparingly the objective functions than Multiobjective Evolutionary Algorithms without models do (typically NSGA-II, \cite{NSGAII}).
	
	As the Pareto set takes up a large part of the design space when many objectives are considered, it may be impossible to compute an accurate approximation to it within the restricted computational budget. Moreover, it may be irrelevant to provide the whole Pareto set because it contains many uninteresting solutions from the DM's point of view.
	
	In the current article, in addition to the GPs, we make use of an aspiration point \cite{wierzbicki1999reference}. This aspiration point can be implicitly defined as the center of the Pareto front \cite{gaudrie2018budgeted} or a neutral point \cite{wierzbicki1999reference} or an extension of it \cite{zeleny1976theory,buchanan03}. Alternatively, it can be given by the DM as a level to attain, and if possible, to improve. In terms of problem class, we consider any objective functions that yields a bounded Pareto front but no assumptions of convexity or continuity of the Pareto front is needed \cite{pardalos2017non}.
	
	The contributions of this work are twofolds. 
	First, in Section \ref{sec:sequential}, we tailor a classical infill criterion used in Bayesian optimization to intensify the search towards the aspiration point. The new criterion is called mEI. 
	The mEI criterion is one of the few Bayesian criteria with \cite{feliotEWHI2019} and \cite{TEHI} that incorporates user preferences. In fact, it can be seen as a very particular instance of both criteria: it is equivalent to a Truncated EHI \cite{TEHI} with infinite lower bounds, and it is equivalent to a Weighted EHI \cite{feliotEWHI2019} with an indicator weighting function of a part of the objective space. But, as we will explain in Sections \ref{sec:mEI} and \ref{sec-center}, the mEI is much simpler to tune and compute.
	
	Second, in Section \ref{sec:parallel}, we propose and study a multi-point extension to this targeting criterion, named q-mEI. We also explain why a tempting alternative criterion to q-mEI is inappropriate for optimization.
	Numerical tests comparing the sequential and batch versions of the mEI algorithm with the Bayesian EHI and the well-known NSGA-II algorithms are reported.
	
	\section{Bayesian multi-objective optimization with sequential targeting}
	\label{sec:sequential}
	
	\subsection{mEI: a new infill criterion for targeting parts of the objective space}
	\label{sec:mEI}
\begin{figure}[h!]
		\centering
		\includegraphics[width=0.32\textwidth]{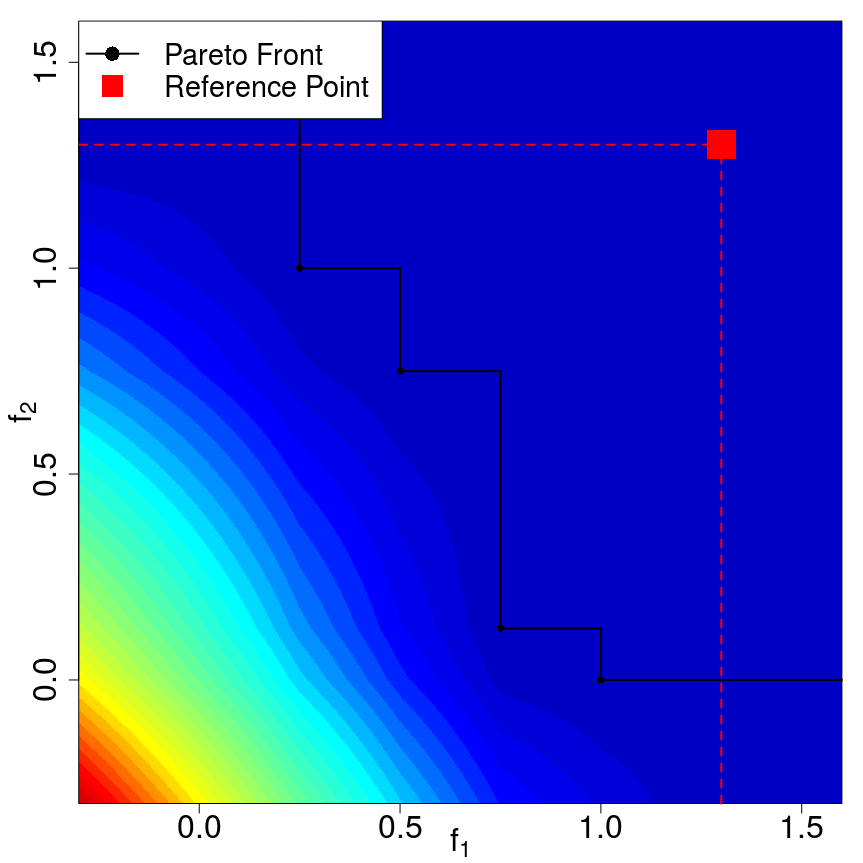}
		\includegraphics[width=0.32\textwidth]{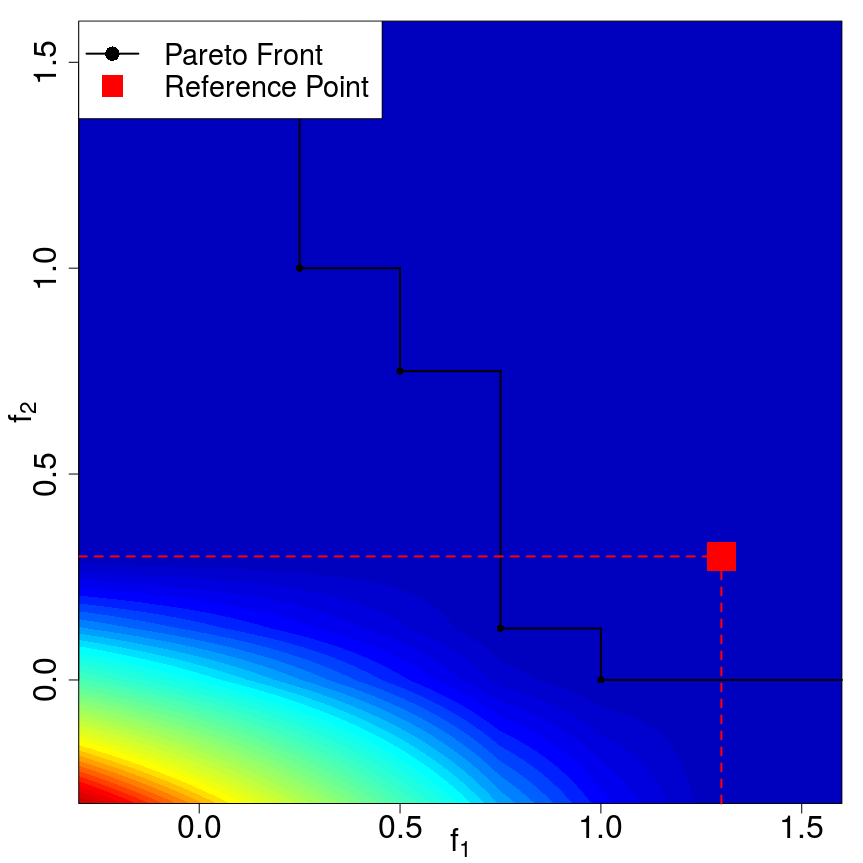}
		\includegraphics[width=0.32\textwidth]{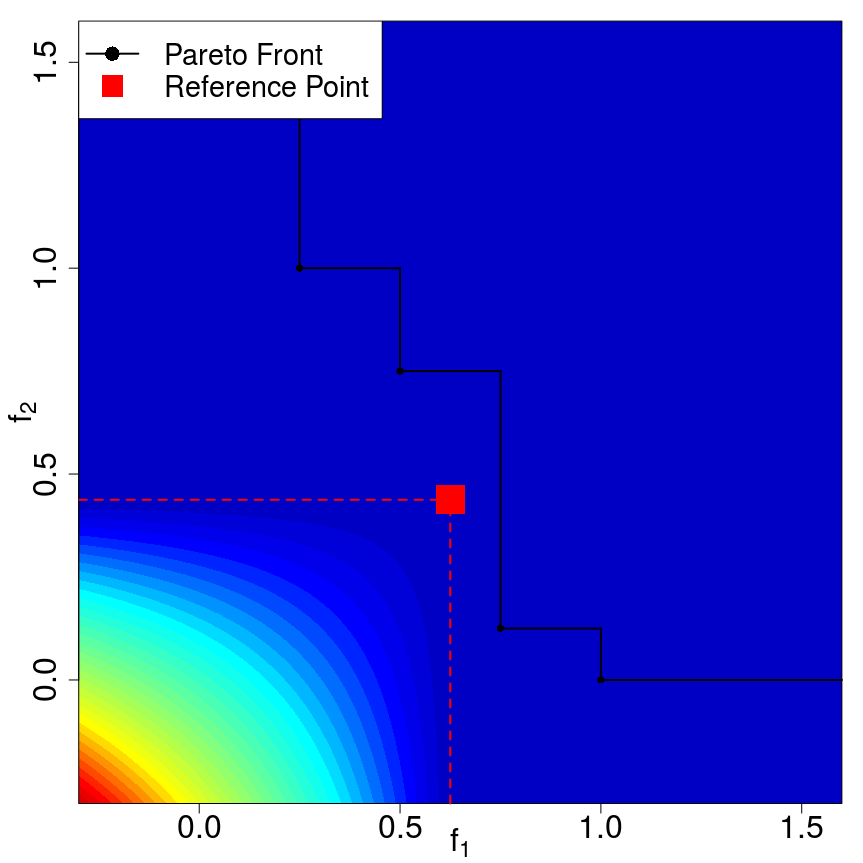}\\
		\includegraphics[width=0.32\textwidth]{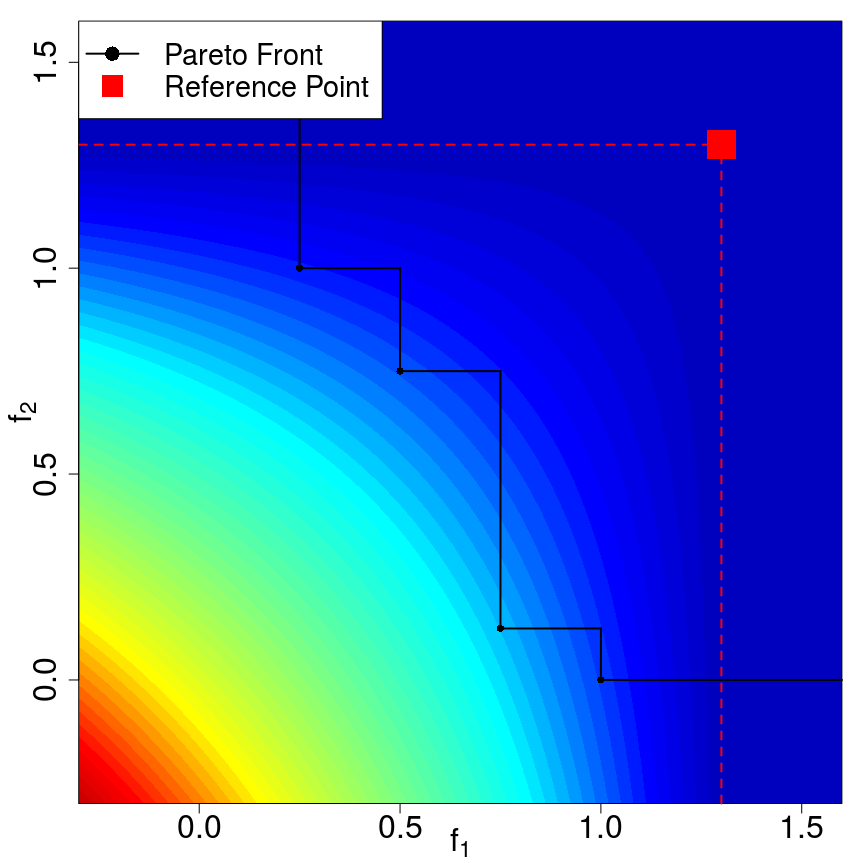}
		\includegraphics[width=0.32\textwidth]{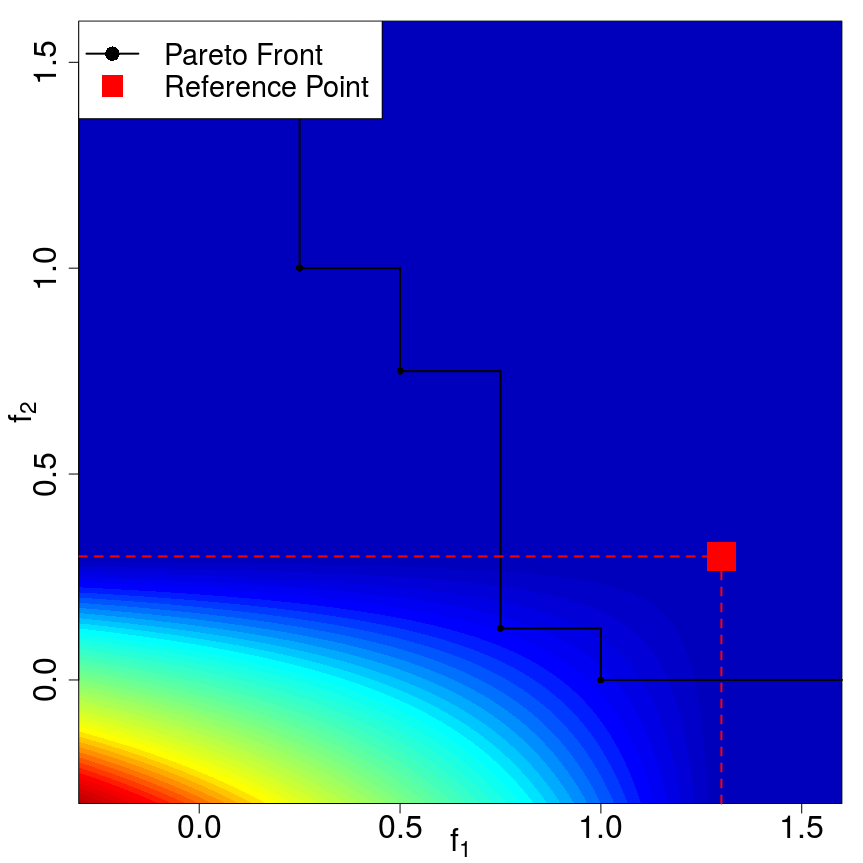}
		\includegraphics[width=0.32\textwidth]{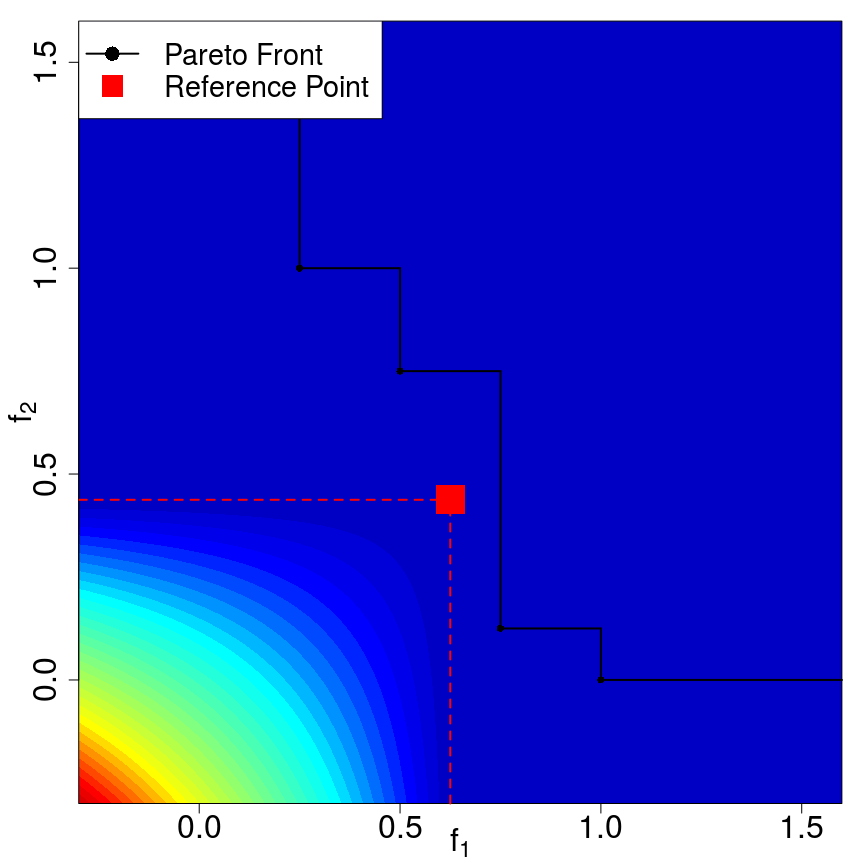}
		\caption{Comparison of the hypervolume improvement (top) and product of objective-wise improvements (bottom) with two objectives. 
			EHI is the expected value of the hypervolume improvement and mEI is the expected value of the product of improvements over $\mathbf Y(\cdot)$ samples. 
			On the left, the reference point ($\RR$, red square) is dominated by the whole front.
			Contrarily to the product of improvements, the hypervolume improvement takes the empirical Pareto front (black) into account.
			In the middle, $\RR$ is only dominated by two points. As both infill criteria equal zero when $f_2>0.3$, using this $\RR$ will promote solutions with small $f_2$ values. On the right, the reference point is non-dominated and the two improvement functions are the same.}
		\label{fig:comparaison_mEI_EHI}
	\end{figure}

	Articulating preferences has already been addressed in multi-objective optimization, see for instance \cite{wierzbicki1980use,book_mcdm,triantaphyllou2000multi,branke2008multiobjective,pref2,deb2006reference}. 
	In Bayesian multi-objective optimization fitted to costly objectives, new points are sequentially added by maximizing an infill criterion whose purpose is to guide the search towards the Pareto set. At each iteration $t$, a new point $\x^{(t+1)}$ is selected and evaluated. $\x^{(t+1)}$ and $\mathbf f(\x^{(t+1)})$ are used to update the metamodels $Y_j(\cdot)$. The Expected Hypervolume Improvement (EHI, \cite{EHI}) is a commonly employed multi-objective infill criterion. It chooses the input which maximizes the expected growth of the hypervolume dominated by $\widehat{\PY}$ up to a reference point $\RR$: $\x^{(t+1)}=\underset{\x\in X}{\arg\max}\text{ EHI}(\x;\RR)$, with 
	\[\text{EHI}(\x;\RR)=\mathbb E[H(\widehat{\PY}\cup\mathbf \{\mathbf Y(\x)\};\RR)-H(\widehat{\PY};\RR)]\]
	where
	\[H(\mathcal A;\RR)=Vol(\bigcup_{\mathbf y\in\mathcal A}\{\mathbf z : \mathbf y\preceq\mathbf z\preceq\RR\})\]
	is the hypervolume indicator of the set $\mathcal A$ up to $\RR$ \cite{TheseZitzler}.
	Classically, $\RR$ is taken beyond the observed Nadir, e.g. \cite{SMS}, in order to cover the entire front. 
	As shown in the first row of Figure \ref{fig:comparaison_mEI_EHI} and already investigated in \cite{auger2009theory,REMOA,auger2012hypervolume,TheseFeliot}, the choice of $\RR$ has a great impact: the farthest from the (empirical) Pareto front, the more the edges are emphasized.
	
	Our approach targets specific parts of the Pareto front first by controlling the reference point $\RR$. Indeed, the choice of $\RR$ is instrumental in deciding the combination of objectives for which \textit{improvement} occurs: $\mathcal I_{\RR}:=\{\y\in \R^m : \y\preceq\RR\}$. Therefore $\RR$ will be set at the aspiration point which can either be provided by the user or the Pareto front center (see next Section) is chosen as default.
	Second, we introduce the mEI (for multiplied Expected Improvements) criterion,
	\begin{equation}
	\text{mEI}(\mathbf x;\RR):=\prod_{j=1}^{m}\text{EI}_j(\mathbf x;R_j)~,
	\label{eq:mEI}
	\end{equation}
	\begin{equation}
	\text{where EI}_j(\mathbf x;R_j)=\E[(R_j-Y_j(\x))_+]
	\label{eq:EI}
	\end{equation}
	is the Expected Improvement below the threshold $R_j$ of the GP $Y_j(\cdot)$.
	Here, $(\cdot)_+$ denotes the positive part.
	
	A large part of the motivation for using mEI is that it is an efficient proxy for the EHI criterion and that it is naturally designed for promoting $\mathcal I_\RR$.
	When the objectives are modeled by independent GPs and the reference point is not dominated by the empirical front, $\widehat{\PY}\npreceq\RR$ (in the sense that no vector in $\widehat{\PY}$ dominates $\RR$), one has EHI$(\cdot;\RR)=\text{mEI}(\cdot;\RR)$.
	The proof is straightforward and given in \cite{gaudrie2018budgeted}. 
	mEI is particularly appealing from a computational point of view. 
	Indeed, EHI requires the computation of non-rectangular $m$-dimensional hypervolumes. 
	Even though the development of efficient algorithms for computing the criterion to temper its computational burden is an active field of research \cite{while2012fast,couckuyt2014fast}, especially in bi-objective \cite{EHI,emmerich2016multicriteria} and in three-objective problems \cite{yang2017computing}, the complexity grows exponentially with the number of objectives and of non-dominated points. 
	When $m>3$, expensive Monte-Carlo estimations are required to compute the EHI. An analytic expression of its gradient has been discovered recently and is limited to the bi-objective cases \cite{yang2019multi}.
	On the contrary, mEI and its gradient have a closed-form expression for any number of objectives, and the complexity grows only linearly in $m$ and is independent of the number of non-dominated solutions. Thus, the mEI criterion can be efficiently maximized.
	
	Figure \ref{fig:comparaison_mEI_EHI} compares the hypervolume improvement and product of objective-wise improvements functions, whose expected values correspond to EHI and mEI, respectively, in a two-dimensional objective space. 
	Contrarily to mEI, EHI takes the Pareto front into account.
	Both criteria are equivalent when $\RR$ is non-dominated. 
	
	In our method, while mEI is a simple criterion, the emphasis is put on the management of the reference point. 
	From now on, $\RR$ will be the initial reference point and $\widehat\RR$ its update that controls the next iterates through
	$\x^{(t+1)}=\underset{\x\in X}{\arg\max}\text{ mEI}(\x;\widehat\RR)$.
	$\RR$ expresses the initial goal of the search. Two situations occur. 
	Either this goal can be reached, i.e. there are points of the true Pareto front that also belong to the improvement cone ${\mathcal I}_\RR$, 
	in which case we want to find any of these performance points as fast as possible. 
	Or the initial goal is too ambitious, no point of the Pareto front dominates $\RR$, in which case we set a new achievable goal and try to reach it rapidly: the updated goal is taken as the point belonging to the $\RR$-Nadir line that is the closest to the true Pareto front; this goal is the intersection of $\RR \mathbf N$ with $\PY$ if it exists. If no initial $\RR$ is given, the default goal is to find the center of the Pareto front which is the subject of the next Section.
	
\begin{figure}[h!]
		\centering
		\includegraphics[width=0.8\textwidth]{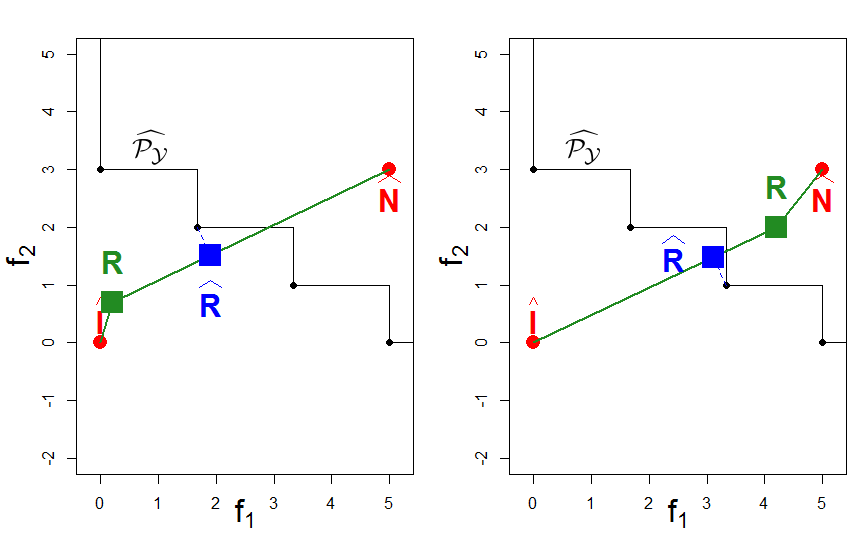}
		\caption{To adapt to $\widehat\PY$, the user-supplied $\RR$ is updated to $\widehat\RR$. 
			Left: $\RR$ is too optimistic, it dominates some of the points of $\widehat\PY$, and $\widehat\RR$ is the closest orthogonal projection of a non-dominated point onto $\RR {\widehat {\mathbf N}}$. Right: the user-provided target has been attained and a more ambitious $\widehat\RR$ is used instead, the closest orthogonal projection of a point of $\widehat\PY$ onto $\widehat{\mathbf I} \RR$.}
		\label{fig:update}
	\end{figure}

	$\widehat\RR$ is controlled to achieve this goal which means avoiding the two pitfalls of global optimization: too much search intensification in already sampled regions, and too much exploration of low potential regions. Excessive intensification is associated with $\widehat\RR$ dominated by already sampled points while superfluous exploration comes from a too ambitious $\widehat\RR$. 
	A compromise is to take $\widehat\RR$ as illustrated in Figure \ref{fig:update}: 
	if $\RR$ dominates at least a point of the empirical Pareto front, $\widehat\RR$ is the point of the $\RR$-estimated Nadir line that is the closest in Euclidean distance to a point of the empirical Pareto front;
	vice versa, if $\RR$ is dominated by at least one calculated point, $\widehat\RR$ is the point of the estimated Ideal-$\RR$ line that is the closest to the empirical Pareto front; finally, in more general cases where $\RR$ is non-dominated, $\widehat\RR$ is set at the point of the broken line joining $\widehat{\mathbf I}$, $\RR$ and $\widehat{\mathbf N}$ that is the closest to $\widehat\PY$.
	In the rare cases where $\widehat\RR$ is dominated after the projection, it is moved on the $\widehat{\mathbf I} \RR \widehat{\mathbf N}$ segments towards $\widehat{\mathbf I}$ until it becomes non dominated.
	By construction $\widehat\RR$ is non-dominated which has a theoretical advantage: mEI$(\x;\widehat\RR)$ is equivalent to EHI$(\x;\widehat\RR)$.
	Thus, when comparing mEI to EHI, note that the complexity of accounting for the empirical Pareto front is carried over from the criterion calculation to the location of the reference point.
	
	Targeting a particular part of the Pareto front leads to a fast local convergence. Once $\widehat\RR$ is on the real Pareto front, the algorithm will dwell on non-improvable values (see left of Figure \ref{fig:comparaison}). 
	To avoid wasting costly evaluations, the convergence has to be checked. To this aim, we estimate $p(\y)$, the probability of dominating the objective vector $\y$, simulating Pareto fronts through conditional GPs. Like the Vorob'ev deviation \cite{molchanov2005theory} used in \cite{binois2015quantifying}, $p(\y)(1-p(\y))$ is a measure of domination uncertainty, which tends to 0 as $p(\y)$ tends to 0 or 1. We assume local convergence and stop the algorithm when the \emph{line-uncertainty}, $\int_{\mathcal L}p(\y)(1-p(\y))d\y$, is small enough, where $\mathcal L$ is the broken line going from $\widehat{\mathbf I}$ to $\RR$ and $\widehat{\mathbf N}$, a line which crosses the empirical front. 
	The convergence detection is described more thoroughly in \cite{gaudrie2018budgeted}.
	A flow chart of this Bayesian targeting search is given in Algorithm \ref{algo}.
	In the absence of preferences expressed through $\RR$, our default implementation uses the center of the front as target, $\widehat\RR \equiv {\mathbf C}$, which is the subject of the next Section.

	\subsection{Well-balanced solutions: the center of the Pareto front}
	\label{sec-center}
	In the same vein as \cite{branke2004finding} where the authors implicitly prefer knee points, we direct the search towards ``well-balanced'' solutions in the absence of explicitly provided preferences. Well-balanced solutions belong to the central part of the Pareto front, defined in the following paragraph, and have equilibrated trade-offs between the objectives.
	
	\paragraph{Definition.}We define the center, $\mathbf C$, of the Pareto front as the projection (intersection in case of a continuous front) of the closest non-dominated point on the Ideal-Nadir line (in the Euclidean objective space).
	An example of Pareto front center can be seen in Figure \ref{fig:refpoints}. This center corresponds visually to an equilibrium among all objectives and is known in game theory as the Kalai-Smordinsky solution with a disagreement point at the Nadir \cite{KS}. 
	Using $\mathbf C$ as default preference in Algorithm \ref{algo} is equivalent to providing any $\RR$ on the line between $\mathbf I$ and $\mathbf N$. In this case, the aspiration level before updating is the Ideal point, which is reasonable in the absence of other preference. 
	The Pareto front center has the property of being insensitive to a linear scaling of the objectives in a bi-objective case\footnote{
		Non-sensitivity to a linear scaling of the objectives is true when the Pareto front intersects the Ideal-Nadir line. Without intersection, exceptions may occur for $m \ge 3$.
	}. $\mathbf C$ has also a low sensitivity to perturbations of the Ideal or the Nadir point: under mild regularity conditions on the Pareto front, $\vert\frac{\partial C_i}{\partial I_j}\vert$ and $\vert\frac{\partial C_i}{\partial N_j}\vert<1$, $i,j=1,\dotsc,m$.
	
	\paragraph{Estimation.}As the Ideal and the Nadir of the empirical Pareto front will sometimes be weak substitutes for the real ones (leading to a biased estimated center), those two points have to be truly estimated. 
	The probabilistic nature of the metamodels (GPs) allows to simulate possible responses of the objective functions. Conditional GP simulations are thus performed to create possible Pareto fronts, each of which defines a sample for $\mathbf I$ and $\mathbf N$. The estimated Ideal and Nadir are the medians of the samples. 
	The intersection between the line $\mathcal{\widehat{L}}$ joining those points and the empirical Pareto front (or the projection if there is no intersection) is the estimated center $\widehat{\mathbf C}$. 
	The same estimation methodology is applied when $\RR$ is provided to estimate $\mathbf I$ and $\mathbf N$ for computing $\widehat\RR$.\vskip\baselineskip
	
	Other properties concerning the center of the Pareto front, and further details regarding the estimations of $\mathbf I$ and $\mathbf N$'s are given in \cite{gaudrie2018budgeted}.\vskip\baselineskip
	
	\begin{algorithm}[H]
		\KwData{
			Create and evaluate an initial DoE of $n$ designs;\\Initialize $m$ GPs $Y_j(\cdot)$ for each objective $f_j(\cdot), j=1,\dotsc,m$;\\
			$t=n$; \emph{line-uncertainty}=$+\infty$; budget;}
		\While{(line-uncertainty$>\varepsilon$) \textbf{and} ($t\le$budget) }{
			{Estimate the Ideal and Nadir, $\widehat{\mathbf I}$ and $\widehat{\mathbf N}$\;}
			\uIf{$\RR$ given \tcc*[r]{adapt $\widehat{\RR}$ to the current Pareto front}}  {Compute $\widehat\RR$ as the closest point from the broken line joining $\widehat{\mathbf I}$, $\RR$ and $\widehat{\mathbf N}$ to $\widehat\PY$\;}
			\Else{
				\tcc{no $\mathbf R$ given, default to center}
				{Estimate the center of the Pareto front $\widehat{\mathbf C}$, $\widehat{\mathbf R}=\widehat{\mathbf C}$\;}
			}
			{$\x^{(t+1)}=\underset{\x\in X}{\arg\max}\text{ mEI}(\x;\widehat{\mathbf R})$\;}
			{evaluate $f_j(\x^{(t+1)})$, update the GPs and $\widehat\PY$\;}
			{compute \textit{line-uncertainty}($\mathbf Y(\cdot)$, $\widehat{\mathbf I}$, $\widehat{\mathbf N}$)\;}
			{$t=t+1$\;}
		}
		\caption{The $\RR / \mathbf{C}$-mEI Bayesian targeting Algorithm
		}
		\label{algo}
	\end{algorithm}
	
	\subsection{First illustrations: targeting with the mEI criterion}
	\label{sec:experiment_targeting_center_mei}
\begin{figure}[h!]
		\centering
		\includegraphics[width=\textwidth]{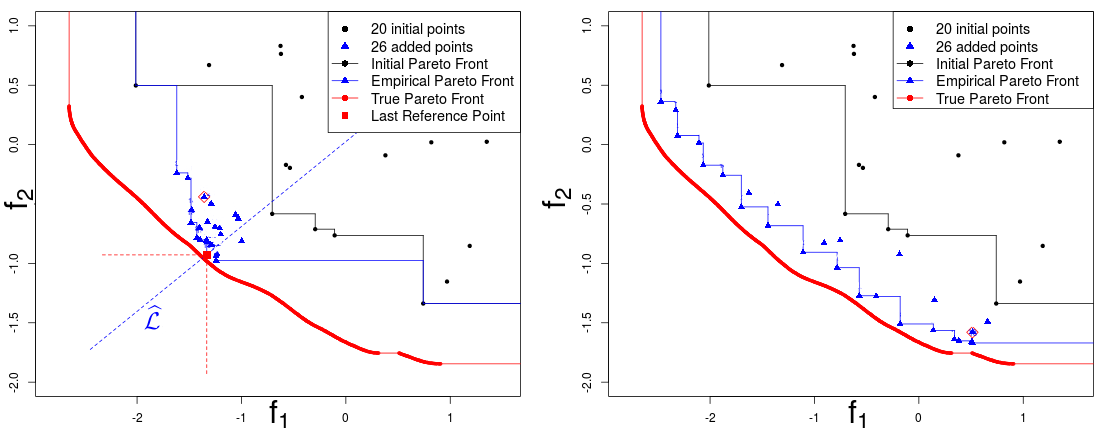}
		\caption{Bi-objective optimization with the $\mathbf C$-mEI algorithm (left). The initial approximation (black) has mainly been improved around the center. Compared with a standard EHI (right), the proposed methodology achieves convergence to the central part of the front. EHI considers more compromises between objectives, but cannot converge within the given budget (26 evaluations).}
		\label{fig:comparaison}
	\end{figure}

\begin{figure}[h!]
		\centering
		\includegraphics[width=0.5\textwidth]{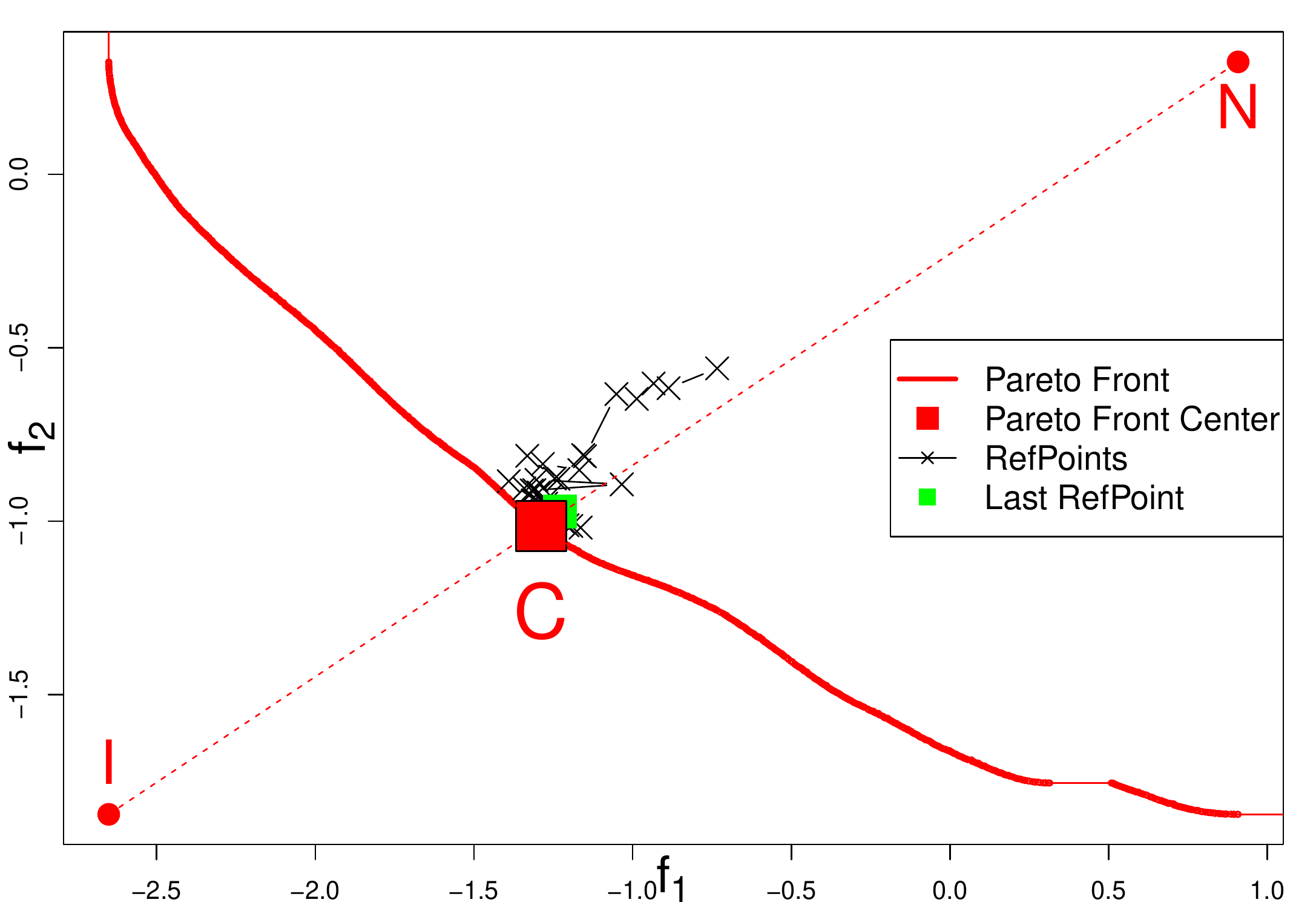}
		\caption{Reference points $\RR$ successively used for directing the search during the $\mathbf C$-mEI run of Figure \ref{fig:comparaison}. They lie close to the dashed Ideal-Nadir line ($\mathbf I\mathbf N$) and lead the algorithm to the center of the Pareto front $(\mathbf C)$.}
		\label{fig:refpoints}
	\end{figure}

	We apply the proposed methodology to the MetaNACA benchmark which is built from real-world airfoil aerodynamic data as described in \cite{gaudrie2018budgeted}. 
	The chosen version of the problem has $d=8$ dimensions and $m=2$ objectives, the negative lift and the drag, to be minimized.
	First, when no preferences are given, the center of the Pareto front is targeted.
	Figure \ref{fig:comparaison} shows that, compared with standard techniques, the proposed methodology leads to a faster and a more precise convergence to the central part of the Pareto front at the cost of a narrower covering of the front. The results are plotted at the iteration which triggers the convergence criterion: only marginal gains would indeed be obtained by continuing to target the same region. Figure \ref{fig:refpoints} indicates how $\RR$ evolves to direct the search to the true center of the Pareto front, $\mathbf C$.
	
\begin{figure}[h!]
		\centering
		\includegraphics[width=0.8\textwidth]{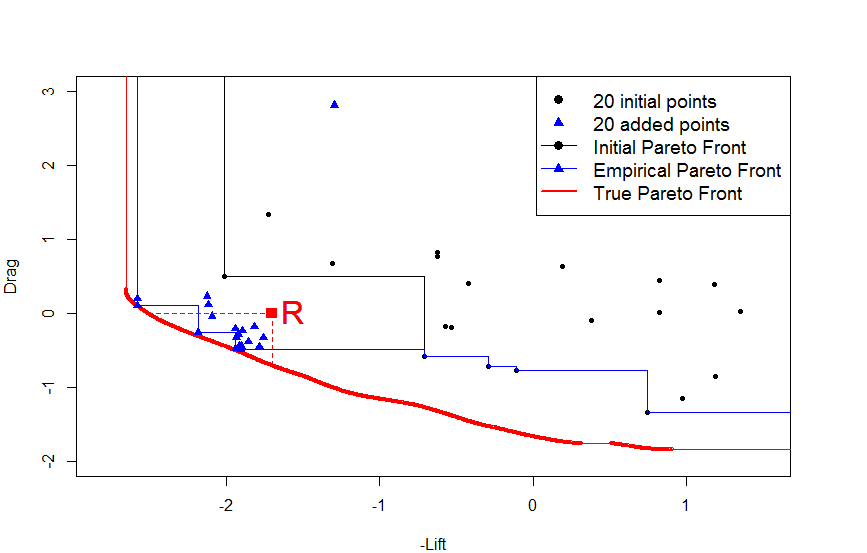}
		\caption{Optimization run targeting an off-centered part of the Pareto front through $\RR$. After 20 iterations, the Pareto front approximation has been improved in the left part, as specified by $\RR$. The successive reference points $\widehat{\RR}$ used by mEI are shown in Figure \ref{fig:refpoints_targeting}.}
		\label{fig:not_centered2}
	\end{figure}

\begin{figure}[h!]
		\centering
		\includegraphics[width=0.6\textwidth]{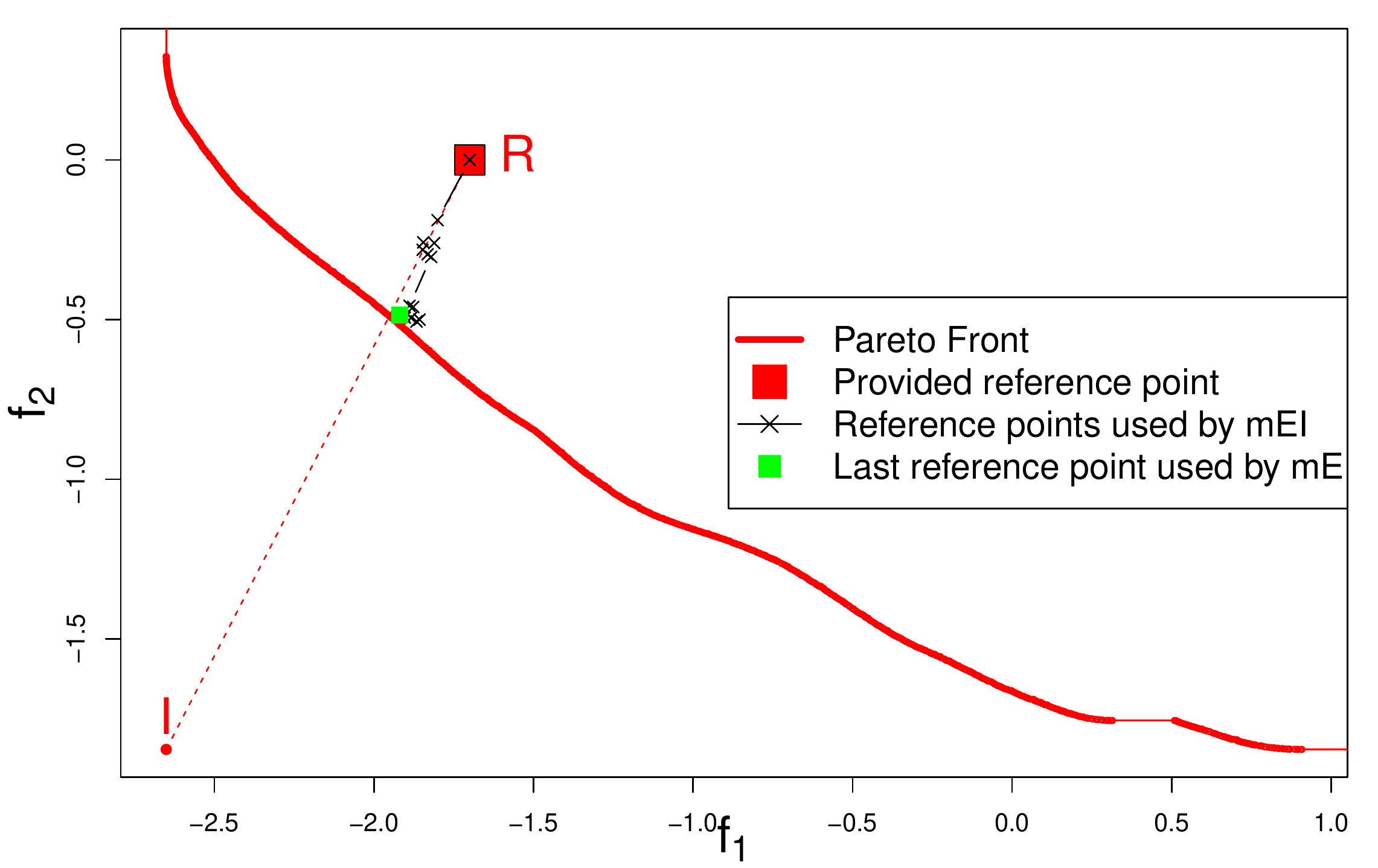}
		\caption{Reference points $\widehat\RR$ successively used for directing the search during the run of Figure \ref{fig:not_centered2}, where $\RR=(-1.7,0)^\top$ is provided. $\widehat\RR$ adjusts to the current approximation front to direct the algorithm in a region of the Pareto front that dominates $\RR$.}
		\label{fig:refpoints_targeting}
	\end{figure}

	Now, we provide the reference point $\mathbf R=(-1.7,0)^\top$ to explicitly target the associated region $\mathcal I_\RR$. A sample convergence of the $\RR$-mEI algorithm is shown in Figure \ref{fig:not_centered2} through the sampled $\mathbf f(\x^{(t)})$ and Figure \ref{fig:refpoints_targeting} gives the associated updated reference points $\widehat\RR$.
	It is seen that mEI($\cdot,\widehat\RR$) effectively guides the search towards the region of progress over $\RR$. Upon closer inspection, it is seen that the points are not spread within $\mathcal I_\RR$ as they would be with EHI($\cdot,\RR$) as the mEI criterion targets a single point on the Pareto front.
	
	
	\section{Targeted Bayesian multi-objective optimization by batch}
	\label{sec:parallel}
	In the context of costly objective functions, the temporal efficiency of Bayesian optimization algorithms can be improved by evaluating the functions in parallel on different computers (or on different cluster nodes). 
	A batch version of these Bayesian algorithms directly stems from replacing the infill criteria with their multi-point pendants: if $q$ points are produced by the maximization of the infill criterion, 
	the $\mathbf f(\cdot)$'s can then be calculated in parallel. 
	In some cases, there is a side benefit to the multi-point criterion in that it makes the algorithms more robust to inadequacies between the GPs and the true functions by spreading the points at each iteration while still complying with the infill criteria logic. 
	
	In a mono-objective setting, the multi-point Expected Improvement (q-EI) introduced in \cite{TheseSchonlau} searches an optimal batch of $q$ points, instead of looking for only one. In \cite{qEI} it is defined as
	\begin{equation}
	\text{q-EI}(\{\x^{(t+1)},\dotsc,\x^{(t+q)}\})=\mathbb E[\underset{i=1,\dotsc,q}{\max}(f_{min}-Y(\x^{(t+i)}))_+]=\mathbb E[(f_{min}-\underset{i=1,\dotsc,q}{\min}Y(\x^{(t+i)}))_+]\label{eq:qEI}
	\end{equation}
	$\{\x^{(t+1)*},\dotsc,\x^{(t+q)*}\}$ maximizing (\ref{eq:qEI}) are $q$ promising points to evaluate simultaneously.
	It is clear from the q-EI criterion that the price to pay for multi-point infill criteria is an increase in the dimension of the inner optimization loop that creates the next iterates. In Algorithm \ref{algo}, the next iterate $\x^{(t+1)}$ results from an optimization in $d$ dimensions, while in a $q$-points algorithm there are $d\times q$ unknowns.
	
	The multi-point Expected Improvement has received some attention recently, see for instance \cite{ginsbourger2010towards,Asynchrone1,Asynchrone2,janusevskis2012expected,frazier2012parallel}, where the criterion is computed using Monte-Carlo simulations. It has been calculated in closed form for $q=2$ in \cite{qEI} and extended for any $q$ in \cite{qEIFormule}.
	An expression and a proxy for its gradient have then been calculated for efficiently maximizing it in $X^q$ \cite{marmin2015differentiating,marmin2016efficient}.
	\vskip\baselineskip
	
	In the same spirit, we wish to extend the multi-objective mEI criterion so that it returns $q$ points to evaluate in parallel. 
	In \cite{horn2015model}, several techniques have been proposed to obtain a batch of $q$ different locations where the functions of multi-objective problems can be evaluated in parallel. 
	They either rely on the simultaneous execution of multi-objective searches with $q$ different goals (e.g., \cite{deb2006reference}), or on the simultaneous evaluation of $q$ points located on a surrogate of the Pareto front or finally on $q$ sequential steps of a multi-objective kriging believer strategy \cite{qEI}.
	A problem involving q-EI's with two objectives is presented in the Chapter 3 of \cite{ribaud_phd} but the formulation is likely to have the same flaws as the mq-EI below, i.e., each point can optimize only a criterion.
	In the current work, we investigate a multi-objective criterion whose maximization yields $q$ points. The resulting strategy is therefore optimal with respect to the criterion.

	\subsection{A naive and a correct batch versions of the mEI}
	
	mEI being a product of EI's, a first approach to extend the mEI criterion to a batch of $q$ points is to use the product of single-objective q-EI's (called mq-EI for ``multiplicative q-EI'') using $R_j$ instead of $\underset{i=1,\dotsc,t}{\min}f_j(\x^{(i)})$ in (\ref{eq:qEI}):
	\begin{align}
	\text{mq-EI}(\{\x^{(t+1)},\dotsc,\x^{(t+q)}\};\RR) = \prod_{j=1}^{m}\text{q-EI}_j(\{\x^{(t+1)},\dotsc,\x^{(t+q)}\};R_j)\nonumber\\ =\prod_{j=1}^{m}\E[\underset{i=1,\dotsc,q}{\max}(R_j-Y_j(\x^{(t+i)}))_+] =\E[\prod_{j=1}^{m}\underset{i=1,\dotsc,q}{\max}(R_j-Y_j(\x^{(t+i)}))_+ ]
	\label{eq:mqEI}
	\end{align}
	because the $Y_j(.)$'s are assumed independent.
	This criterion has however the drawback of not using a product of joint improvement in all objectives, as the $\max$ among the $q$ points is taken independently for each objective $j$ considered. This may lead to undesirable behaviors: the batch of $q$ optimal points using this criterion may be composed of optimal points w.r.t. each individual EI$_j$. 
	For example with $m=2$ and $q=2$, a batch $\{\x^{(1)*},\x^{(2)*}\}$ with promising $Y_1(\x^{(1)*})$ and $Y_2(\x^{(2)*})$ may be optimal, without taking $Y_2(\x^{(1)*})$ and $Y_1(\x^{(2)*})$ into account.
	$\x^{(1)*}$ and $\x^{(2)*}$ may not even dominate $\RR$ while scoring a high mq-EI. For these reasons, the mq-EI criterion breaks the coupling through $\x$ between the functions, allocating marginally each point to an objective. mq-EI does not tackle multi-objective problems.
	
	Following the definition of q-EI (\ref{eq:qEI}), a proper multi-point extension of mEI (\ref{eq:mEI}) is
	
	\begin{equation}
	\text{q-mEI}(\{\x^{(t+1)},\dotsc,\x^{(t+q)}\};\RR)=\E\left[\underset{i=1,\dotsc,q}{\max}\left(\prod_{j=1}^{m}(R_j-Y_j(\x^{(t+i)}))_+\right)\right]
	\label{eq:qmEI}
	\end{equation}
	
	\subsection{Properties of both criteria}
	\label{sec:prop_qmei_mqei}
	We now give some properties and bounds for both criteria.
	
	\begin{proposition}
		When evaluated twice at the same design, mq-EI and qm-EI reduce to mEI: mq-EI$(\{\x,\x\};\RR)=\text{q-mEI}(\{\x,\x\};\RR)=\text{mEI}(\x;\RR)$.
	\end{proposition}
	\noindent Proof:\\
	mq-EI$(\{\x,\x\};\RR)=\prod_{j=1}^{m}\text{q-EI}_j(\{\x,\x\};R_j)=\prod_{j=1}^{m}$EI$_j(\x;R_j)=\text{mEI}(\x;\RR)$.\\
	q-mEI$(\{\x,\x\};\RR)=\E[(\prod_{j=1}^{m}(R_j-Y_j(\x))_+)]=\text{mEI}(\x;\RR).$ $\square$

	\begin{proposition}
		When $\PY\npreceq\RR$, q-mEI calculated at two training points $\x$ and $\x'$ is null. q-mEI calculated at one training point $\x$ and one new point $\x''$ reduces to mEI at the latter: q-mEI$(\{\x,\x'\};\RR)=0$, q-mEI$(\{\x,\x''\};\RR)=\text{mEI}(\x'';\RR)$.
	\end{proposition}
	
	\noindent Proof:\\
	As $\x$ and $\x'$ are training points, $\mathbf Y(\x)$ and $\mathbf Y(\x')$ are no longer random variables, and the expectation vanishes. Since $\RR$ is not dominated by the observed values $\y=\mathbf Y(\x)$ and $\y'=\mathbf Y(\x')$, $\prod_{j=1}^{m}(R_j-Y_j(\x))_+=\prod_{j=1}^{m}(R_j-y_j)_+=0$ and the same occurs with $\y'$.  Finally, q-mEI$(\{\x,\x'\};\RR)=0$.\\
	In the case of one observed $\x$ and one unobserved $\x''$, $\prod_{j=1}^{m}(R_j-Y_j(\x''))_+\ge\prod_{j=1}^{m}(R_j-Y_j(\x))_+=0$, and q-mEI$(\{\x,\x''\};\RR)=\E[\prod_{j=1}^{m}(R_j-Y_j(\x''))_+]=\text{mEI}(\x'';\RR)$. $\square$\\
	
	Even though these properties seem obvious and mandatory for a multi-point infill criterion, they do not hold for mq-EI. To see this, let us consider a case with $m=2$ objectives, $\RR$ a non-dominated reference point, and $\x$ and $\x'$ two evaluated designs with responses $\y=\mathbf f(\x)=(y_1,y_2)^\top$, $\y'=\mathbf f(\x')=(y_1',y_2')^\top$, satisfying $y_1<R_1<y_1'$ and $y_2'<R_2<y_2$. By definition, mq-EI$(\{\x,\x'\};\RR)=\prod_{j=1}^2\E[\max((R_j-y_j)_+,(R_j-y_j')_+)]=(R_1-y_1)(R_2-y_2')>0$.
	Furthermore, mq-EI$(\{\x,\x''\};\RR)=\prod_{j=1}^2\E[\max((R_j-y_j)_+,(R_j-Y_j(\x''))_+)]=\text{EI}_2(\x'';R_2)\times\E[\max((R_1-y_1),(R_1-Y_1(\x''))_+)]$
	$> \text{EI}_2(\x'';R_2)\times \text{EI}_1(\x'';R_1) = \text{mEI}(\x'';\RR)$.
	
	\vskip\baselineskip
	Some bounds can also be computed. We assume $q \ge m$ which will usually be verified. Let us denote $\x^{(j)*}$ the maximizers of EI$_j(\cdot;R_j)$ for $j=1,\dotsc,m$; $\x^{(m+1)*},\dotsc,\x^{(q)*}$ any other points and $\x^*$ the maximizer of mEI$(\cdot,\RR)$. Then,
	
	\begin{align*}
	\underset{\x^{(1)},\dotsc,\x^{(q)}}{\max}\text{ mq-EI}(\{\x^{(1)},\dotsc,\x^{(q)}\};\RR)=
	\underset{\x^{(1)},\dotsc,\x^{(q)}}{\max}\prod_{j=1}^{m}\text{q-EI}_j(\{\x^{(1)},\dotsc,\x^{(q)}\};\RR_j)\\
	\ge \prod_{j=1}^{m}\text{q-EI}_j(\{\x^{(1)*},\dotsc,\x^{(m)*},\x^{(m+1)*},\dotsc,\x^{(q)*}\};\RR_j)
	\ge\prod_{j=1}^{m}\text{EI}_j(\x^{(j)*};\RR_j)
	\end{align*}
	
	This inequality shows that mq-EI's maximum value is greater than the product of expected improvement maxima, which shows that this criterion does not minimize $f_1,\dotsc,f_m$ jointly.
	The last term can be further lower bounded, $\prod_{j=1}^{m}\text{EI}_j(\x^{(j)*};\RR)\ge\prod_{j=1}^{m}\text{EI}_j(\x^*;\RR)=\text{mEI}(\x^*;\RR)$.\vskip\baselineskip
	
	For q-mEI, a trivial lower bound is the mEI maximum:
	$\underset{\x^{(1)},\dotsc,\x^{(q)}}{\max}\text{q-mEI}(\{\x^{(1)},\dotsc,\x^{(q)}\};\RR)\ge\underset{\x}{\max}\text{ mEI}(\x;\RR)=\text{mEI}(\x^*;\RR)$.\vskip\baselineskip
	
	These lower bounds indicate that more improvement is expected within the $q$ steps than during a single mEI step. 
	
	\subsection{Experiments with the batch targeting criteria}
	\label{sec:parallel_expe}
	We now investigate the capabilities of the batch version of mEI. 
	First, in Section \ref{sec:expemqEI}, a comparison between q-mEI and mq-EI is made on the basis of two simple one-dimensional quadratic functions. This example illustrates why q-mEI is the correct multi-point extension of mEI. 
	Then, in Section \ref{section:center_targeting}, the batch criterion q-mEI is compared with the sequential mEI for finding the Pareto front center using the physically meaningful functions of the MetaNACA test bed \cite{gaudrie2018budgeted}.
	Finally, in Section \ref{section:region_targeting}, a larger comparison is carried out: it involves Bayesian algorithms with the mEI, q-mEI and EHI criteria plus the NSGA-II algorithm; an off-centered preference region is specified; analytical functions standard in the literature make this last test bed.

	Note that in both Sections \ref{section:center_targeting} and \ref{section:region_targeting}, parallel executions of the algorithms are simulated on sequential computers. 
		As usual in Bayesian optimization, we assume that the computation time is mainly taken by the calls to the objective functions and there are sufficient computing resources so that the speed-up is close to $q$. The term ``wall-clock time'' will therefore mean the number of calls to the objective functions divided by the batch size $q$.
	
	The q-mEI and mq-EI criteria of formula (\ref{eq:qmEI}) and (\ref{eq:mqEI}) are calculated by Monte Carlo simulation with $N=10,000$ samples. 
	To be more precise, q-mEI (\ref{eq:qmEI}) at a candidate batch $\{\x^{(t+1)},\dotsc,\x^{(t+q)}\}$ is computed by averaging over $N$ conditional GPs ${\widetilde{Y_j}}^{(k)}$, which leads to the estimator
	\[\widehat{\text{q-mEI}}(\{\x^{(t+1)},\dotsc,\x^{(t+q)}\};\RR)=\frac1N\sum_{k=1}^{N}\left[\underset{i=1,\dotsc,q}{\max}\left(\prod_{j=1}^{m}(R_j-{\widetilde{Y_j}}^{(k)}(\x^{(t+i)}))_+\right)\right]\]
	Because the optimization of the criteria is carried out in a $q \times d$ dimensional space and the gradients are not available, in the experiments the number of iterates evaluated simultaneously is restricted to $q=2$ and 4. 
	
	\subsubsection{Comparison between mq-EI and q-mEI on quadratic functions}
	\label{sec:expemqEI}
\begin{figure}[h!]
		\centering
		\includegraphics[width=0.6\textwidth]{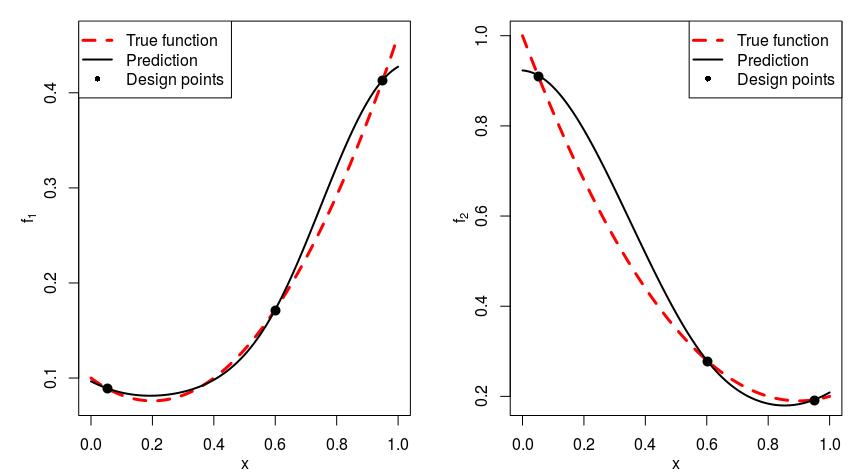}
		\includegraphics[width=0.6\textwidth]{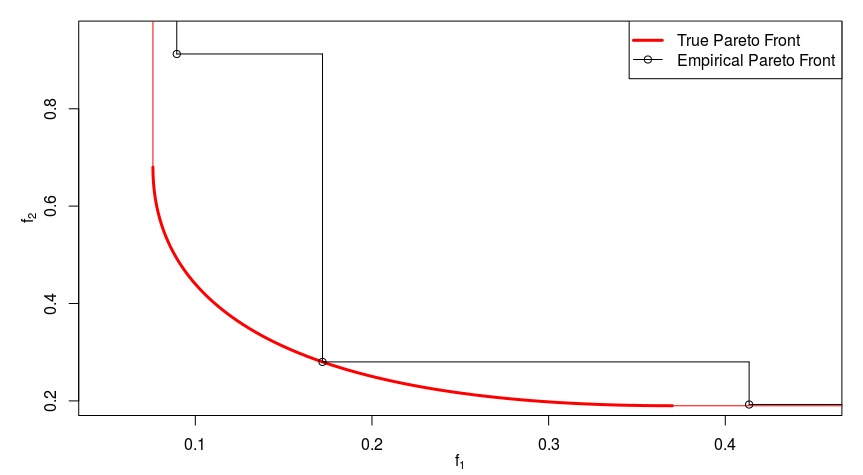}
		\caption{Top: Kriging predictors and true $f_1$ and $f_2$ functions. Bottom: true and empirical Pareto fronts.}
		\label{fig:predict}
	\end{figure}

	To compare q-mEI with mq-EI, we consider a simple example with $d=1$, $q=2$ and $m=2$ quadratic objective functions:
	\[\underset{x \in [0,1]}{\min}(f_1(x),f_2(x))\]
	where $f_1(x)=0.6x^2-0.24x+0.1$ and $f_2(x)=x^2-1.8x+1$, whose minima are respectively 0.2 and 0.9. The multi-objective optimality conditions \cite{miettinen1998nonlinear} show that the Pareto set is $\PX=[0.2,0.9]$ and the Pareto front $\PY=\{\y=(f_1(x),f_2(x))^\top, x\in[0.2,0.9]\}$. $f_1$ and $f_2$ are plotted in red in
	Figure \ref{fig:predict}, both in the design space $X=[0,1]$ and in the objective space.
	Two independent GPs, $Y_1(\cdot)$ and $Y_2(\cdot)$, are fitted to $n=3$ data points, $x^{(1)}=0.05$, $x^{(2)}=0.6$ and $x^{(3)}=0.95$.
	Figure \ref{fig:predict} also shows the kriging predictors $\widehat{f}_1(x)$ and $\widehat{f}_2(x)$, as well as the empirical Pareto front.
	
\begin{figure}[h!]
		\centering
		\includegraphics[width=0.6\textwidth]{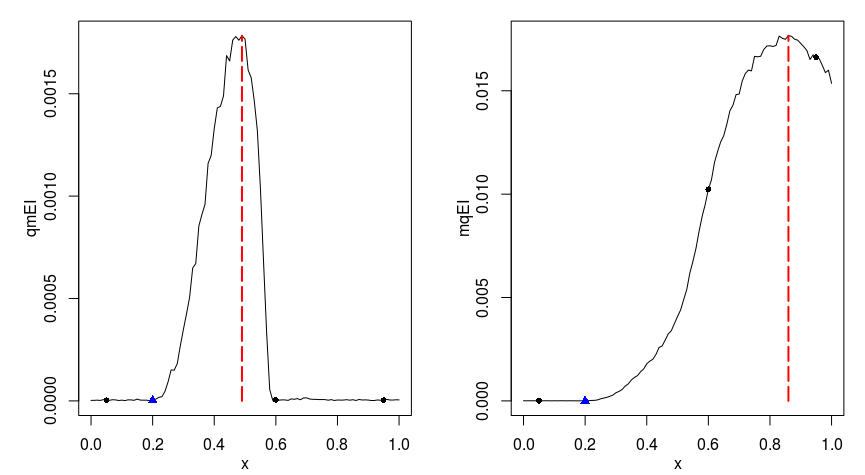}
		\includegraphics[width=0.6\textwidth]{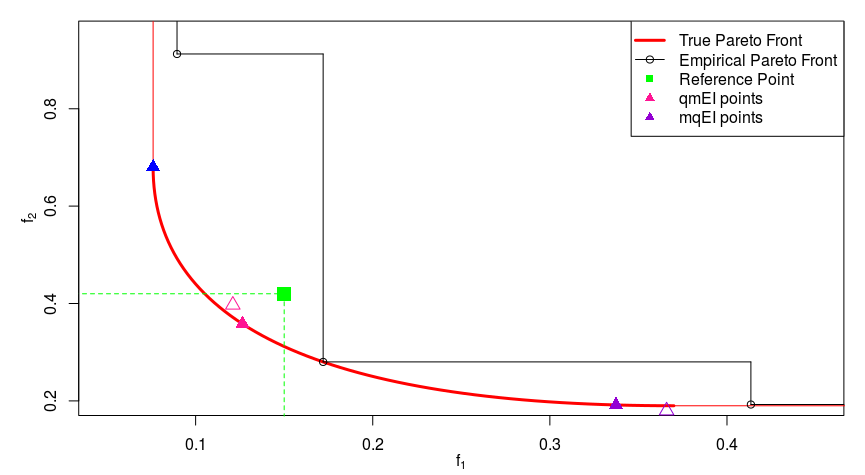}
		\caption{Setting 1: $x^{(n+1)}=0.2$ (blue triangle). Top: q-mEI$(\{x^{(n+1)},x\};\RR)$ (left) and mq-EI$(\{x^{(n+1)},x\};\RR)$ (right) criteria for the second input in the design space. The maximum is achieved at different locations for both criteria. Also notice that for training points $x^{(i)}$ (black dots), mq-EI$(\{x^{(n+1)},x^{(i)}\};\RR)\ne$\text{ mEI}$(x^{(n+1)};\RR)\approx 0$, contrarily to q-mEI$(x^{(n+1)},x^{(i)})$. Bottom: corresponding values for $\mathbf f(x^{(n+2)})$. q-mEI provides an input whose image (pink) dominates $\RR$. On the contrary, mq-EI's solution concentrates on the minimization of the second objective (purple). The transparent triangles correspond to the kriging predictions at $x^{(n+2)}$.}
		\label{fig:first_point_02}
	\end{figure}

	Let us take the non-dominated reference point $\RR=(0.15,0.42)^\top$ that we will use both with mq-EI and q-mEI. With that reference point, shown in green in Figure \ref{fig:first_point_02}, domination of $\RR$ is achieved when $x\in[0.42,0.55]$.
	
\begin{figure}[h!]
		\centering
		\includegraphics[width=0.6\textwidth]{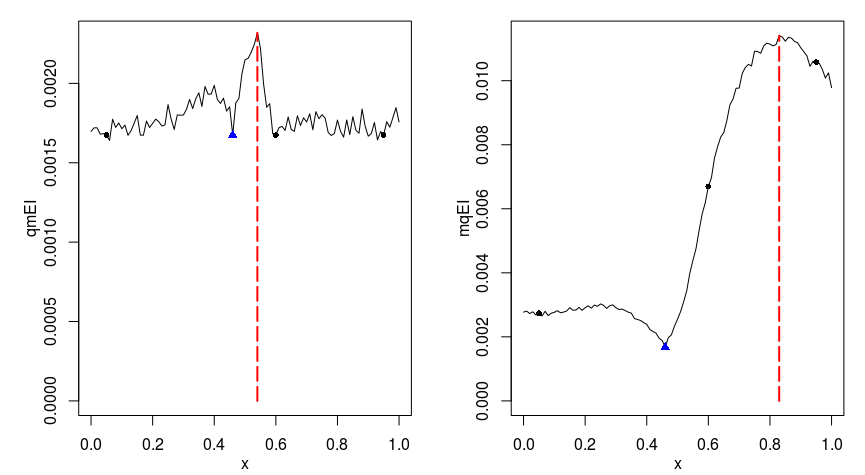}
		\includegraphics[width=0.6\textwidth]{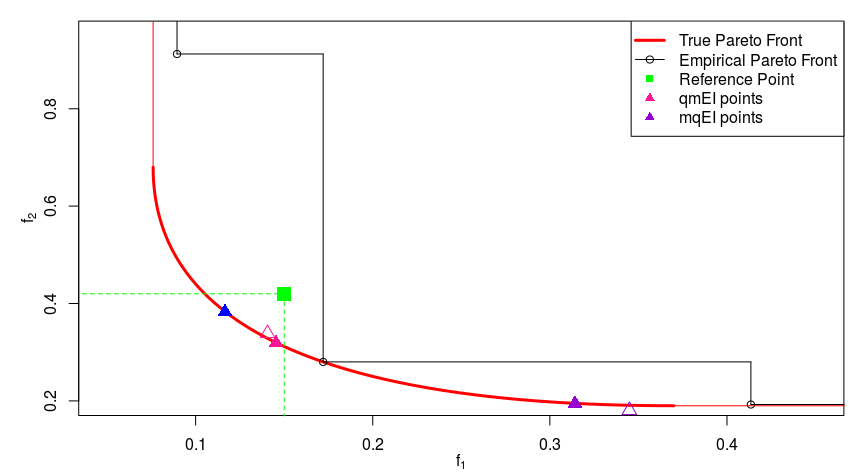}
		\caption{Setting 2: $x^{(n+1)}=0.46$ (blue triangle). Top: q-mEI$(\{x^{(n+1)},x\};\RR)$ (left) and mq-EI$(\{x^{(n+1)},x\};\RR)$ (right) criteria for the second input in the design space whose maximum is again achieved at different locations. Bottom: corresponding values for $\mathbf f(x^{(n+2)})$. q-mEI provides an input whose image (pink) also dominates $\RR$. On the contrary, mq-EI returns an input which concentrates on the minimization of the second objective (purple). The transparent triangles correspond to the kriging predictions at $x^{(n+2)}$.}
		\label{fig:first_point_046}
	\end{figure}

	In a first experiment, we fix $x^{(n+1)}$ (but it is not a training point, its objective values are handled through the GPs) and search for the $x^{(n+2)}$ maximizing mq-EI$(\{x^{(n+1)},x^{(n+2)}\};\RR)$ and q-mEI$(\{x^{(n+1)},x^{(n+2)}\};\RR)$. 
	Besides illustrating the difference between q-mEI and mq-EI, this experiment may serve as an introduction to the asynchronous versions of the batch criteria \cite{janusevskis2012expected} which are important in practical parallel implementations: as soon as one computing node becomes available, the $q$-points criteria are optimized with respect to 1 point while keeping the $q-1$ other points fixed at their currently running values.
	Two different settings are considered whose results are presented in Figures \ref{fig:first_point_02} and \ref{fig:first_point_046}.
	
	In the first setting, $x^{(n+1)}=0.2$ is a bad choice as it corresponds to an extreme point of the Pareto set and its future response will not dominate $\RR$, an information already seen on the GPs. 
	q-mEI gives $x^{(n+2)}=0.49$ which is very close to the (one-step) mEI maximizer,
	hence a relevant input as $\mathbf f(x^{(n+2)})$ will dominate $\RR$. 
	On the contrary, mq-EI separates the objectives. As $x^{(n+1)}$ is a good input for objective $f_1$, the criterion reaches its maximum when $x^{(n+2)}=0.86$, which is a good input when considering $f_2$ alone. Figure \ref{fig:predict} tells us that 0.86 is almost the minimizer of $\widehat{f}_2(x)$. However, the original goal of dominating $\RR$ is not achieved.
	
	In the second setting, $x^{(n+1)}=0.46$ is a good point as its image will dominate $\RR$. 
	q-mEI leads to $x^{(n+2)}=0.53$ whose image also dominates $\RR$. Notice that as 0.46 is chosen for $x^{(n+1)}$, the point that jointly maximizes q-mEI with that first point is slightly larger than 0.48 (the mEI maximizer), and provides more diversity in $\mathcal{I}_\RR$. The second input for maximizing mq-EI is $x^{(n+2)}=0.83$, an input that is good only to minimize $f_2$ (it is almost the same as in the previous case) but $\mathbf f(x^{(n+2)})$ does not dominate $\RR$.
	
\begin{figure}[h!]
		\centering
		\includegraphics[width=0.6\textwidth]{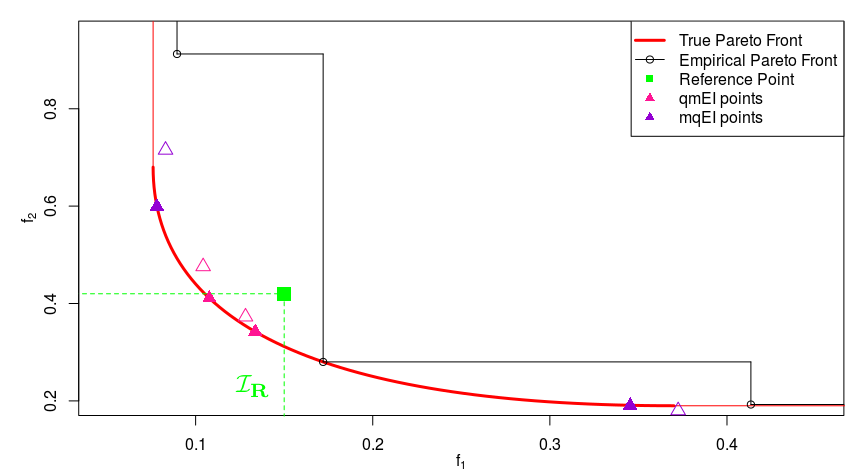}
		\includegraphics[width=0.6\textwidth]{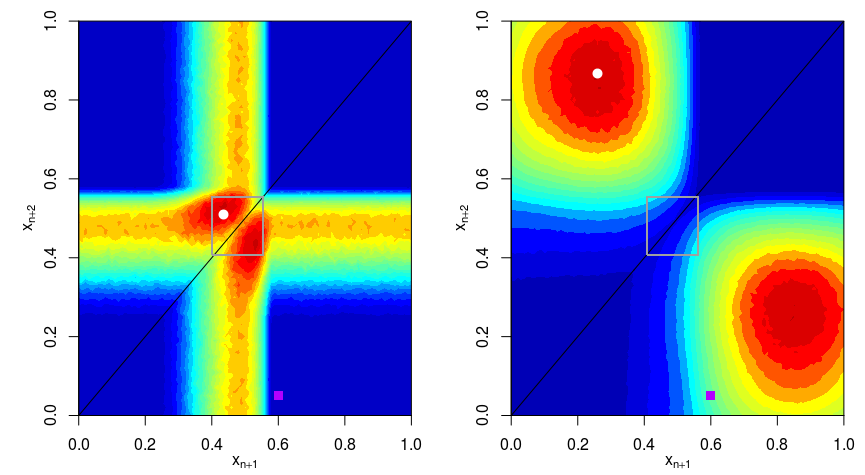}
		\caption{2 points mq-EI and q-mEI. Top: values obtained in the objective space using both criteria. The images of $x^{(n+1)}$ and $x^{(n+2)}$ returned by q-mEI (pink) both dominate $\RR$. None of those returned by mq-EI (purple) are in $\mathcal I_\RR$, they rather improve over each 
			function individually. Transparent triangles correspond to the kriging mean predictor. Bottom: criteria values for varying $(x^{(n+1)}$,$x^{(n+2)})$. For q-mEI (left), the best $x$'s, in dark red, lead to domination of $\RR$ (gray box). Conversely, for mq-EI (right), good $x$'s improve upon $\RR$'s components for each objective. The white dots correspond to both optima. 
			The purple square is an example of a training point pair where q-mEI is null but mq-EI is not (this holds for all other training point pairs that are not shown).} 
		\label{fig:two_points}
	\end{figure}

	Now, we optimize directly mq-EI and q-mEI with respect to both inputs $x^{(n+1)}$ and $x^{(n+2)}$.
	The optimal batches are $\{0.43,0.51\}$ for q-mEI and $\{0.26,0.87\}$ for mq-EI.
	Figure \ref{fig:two_points} shows that these inputs lead to $\mathcal I_\RR$ with q-mEI. On the contrary, the images of mq-EI's optimum are located at the boundaries of the Pareto front and none of them is in $\mathcal I_\RR$.
	Figure \ref{fig:two_points} further indicates that q-mEI is high when both inputs are in the part of the design space that leads to domination of $\RR$ (gray box)
	contrarily to mq-EI, which is high when each input leads to the domination of one component of $\RR$.
	Note that, even though both criteria are symmetric with respect to their $q$ inputs, the symmetry is slightly broken in the Figure because of the Monte-Carlo estimation.
	
	\subsubsection{Batch targeting of the Pareto front center}
	\label{section:center_targeting}
	We now compare the q-mEI with the sequential mEI and the (non targeting) EHI infill criteria. 
	As in Section \ref{sec:experiment_targeting_center_mei}, the tests are performed with the MetaNACA benchmark \cite{gaudrie2018budgeted} in $d=8$ dimensions and with $m=2$ objectives. 
	No user-defined reference point is provided so the center of the Pareto front is targeted. 
	
\begin{figure}[h!]
		\centering
		\includegraphics[width=\textwidth]{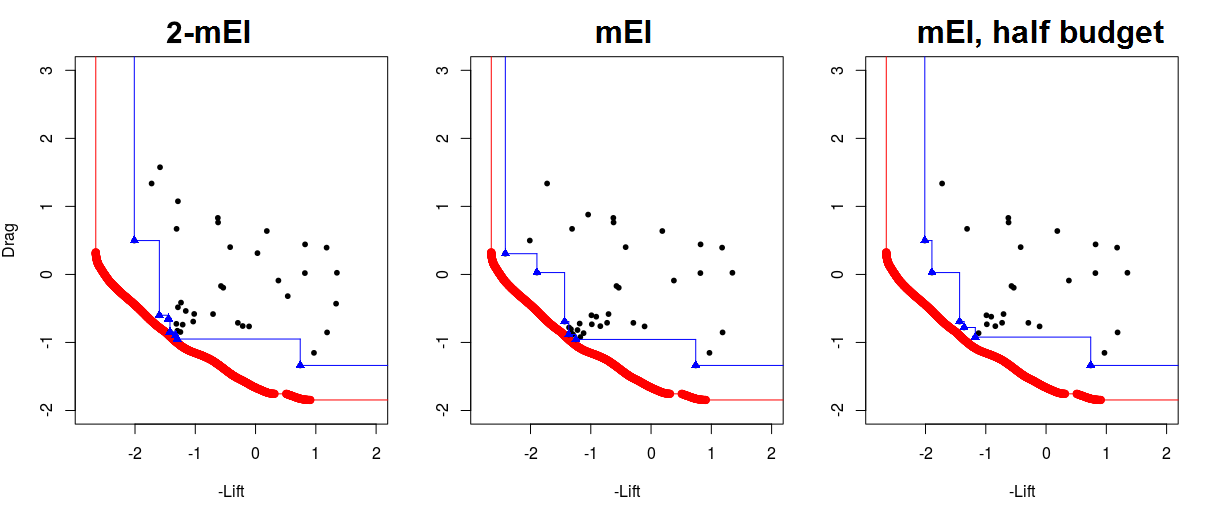}\\
		\includegraphics[width=0.7\textwidth]{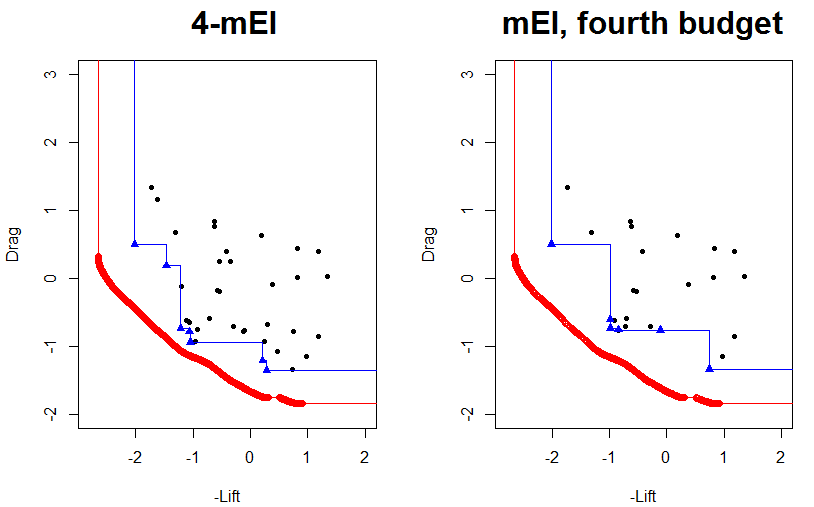}
		\caption{
			Top: example run using 2-mEI (left) with $2\times10$ additional designs, and mEI with 20 (center) and 10 (right) iterations. 
			Bottom: example run using 4-mEI (left) with $4\times5$ additional designs, and mEI with 5 iterations (right). 
			At a fixed wall-clock time, q-mEI converges more accurately to the center than mEI. At a fixed number of evaluations, the degradation is small.
		}
		\label{fig:comparaison_qmEI_mEI}
	\end{figure}

	Figure \ref{fig:comparaison_qmEI_mEI} allows a graphical comparison of the effects of the sequential and batch mEI criteria at constant wall-clock time or constant number of calls to the objective functions.
	Recall that under our assumptions of costly objective functions, 2-mEI with $2\times10$ iterations and mEI with 10 iterations roughly need the same wall-clock time.
	Similarly, 4-mEI with $4\times 5$ iterations and mEI fourth budget take the same time.
	On both rows of the Figure, it is seen that at the same wall-clock time, q-mEI's approximations to the front center (left) are improved when compared to mEI's (right). For an equal number of added points, 2-mEI and mEI provide equivalent approximations to the center, and 4-mEI is slightly degraded (but the time is divided by 4). At the same number of evaluations, the small deterioration of q-mEI's results over those of mEI is explained as follows: when $q$ increases, the batch versions of the criterion affect resources (i.e., choose the $\x$'s to be calculated) with increasingly incomplete information.
	Referring back to the right plot in Figure \ref{fig:comparaison} which corresponds to the same test case, one can remember the type of global yet incomplete convergence that is obtained with the EHI criterion.
	
	Let us now turn to statistically more significant comparisons where runs are repeated. Two performance metrics are used. First, a \emph{restriction of the hypervolume indicator} \cite{TheseZitzler}	to a central part, ${\mathcal I}_w$, of the Pareto front: the hypervolume is computed up to the reference point $\RR_w:=(1-w)\mathbf C+w\mathbf N$, where $\mathbf C$ is the center of the Pareto front, and $\mathbf N$ its Nadir point. $w=0.1$ means that the hypervolume is calculated only for the points that are 
	in a small vicinity\footnote{In case of a linear Pareto front, $\mathcal I_{0.1}$ corresponds to the 10\% most central solutions} of $\mathbf C$.
	Figures showing the part of $\PY$ to which the indicator is restricted can be found in \cite{gaudrie2018budgeted}.
	The second performance metric is the \emph{time to target}.
	It corresponds to the number of function evaluations taken by an algorithm to dominate a user-defined reference point. 
	Here, $\RR_w$ serves as reference.
	
	Tables \ref{tab:comparaison_qmei_mei} and \ref{tab:comparaison_qmei_mei_time} report the averages and standard deviations of the performance metrics, the restricted hypervolume and the time to target, calculated over 10 independent runs.
	The columns ``mEI'' and ``q-mEI'' correspond to optimizations using the same number of evaluations, i.e., 20 and 50 for $d=8$ and 22, respectively. 
	The column ``mEI, half budget'' corresponds to runs stopped after 10 and 25 iterations for $d=8$ and 22, hence with the same wall-clock time as 2-mEI.  The same explanation holds for the ``EHI'' and ``EHI, half budget'' columns. 
	
	\begin{table}[h!]
		\centering
		\caption{Hypervolume indicator computed in ${\mathcal I}_{0.1}$ for 2-mEI, mEI and EHI at identical number of evaluations and wall-clock times. 
			Averages \textit{(std. deviation)} over 10 runs. 
		}
		\begin{tabular}{|c|c|c|c|c|c|}
			\hline
			& 2-mEI & mEI & mEI, half budget & EHI & EHI, half budget\\\hline
			$d=8$ & 0.087 \textit{(0.09)} & 0.256 \textit{(0.09)} & 0.134 \textit{(0.15)} & 0.025 \textit{(0.04)} & 0\\
			$d=22$ & 0.128 \textit{(0.08)} & 0.222 \textit{(0.12)} & 0.139 \textit{(0.10)} & 0.153 \textit{(0.09)} & 0.079 \textit{(0.08)}\\\hline
		\end{tabular}
		\label{tab:comparaison_qmei_mei}
	\end{table}
	
	\begin{table}[h!]
		\centering
		\caption{Number of function evaluations required to dominate ${\mathbf R}_{0.1}$ for 2-mEI, mEI and EHI. Averages \textit{(std. deviation)} over 10 runs. For 2-mEI, the wall-clock time is the number of evaluations divided by $q=2$, while it is the same number for mEI and EHI.
		}
		\begin{tabular}{|c|c|c|c|}
			\hline
			& 2-mEI & mEI & EHI\\\hline
			$d=8$ & 14 \textit{(5.6)} & 8.4 \textit{(5.4)} & 26.8 \textit{(2.6)}\\
			$d=22$ & 6.4 \textit{(6.4)} & 6.3 \textit{(7.2)} & 21.4 \textit{(13.9)}\\\hline
		\end{tabular}
		\label{tab:comparaison_qmei_mei_time}
	\end{table}
	
	These empirical results indicate that at the same wall-clock time, q-mEI outperforms mEI in attainment time of $\RR_{0.1}$ (Table \ref{tab:comparaison_qmei_mei_time}): even though mEI attains this central target after less function evaluations, q-mEI is able to perform $q$ (2 here) calls to $f(\cdot)$ during one iteration. The center of the Pareto front is therefore attained faster in wall-clock time with q-mEI than with mEI. Both criteria widely outperform EHI which, again, attempts to uncover the whole Pareto front but does not get as close to the true Pareto front's center. Notice that at half the computational budget (that is to say, same wall-clock time than q-mEI), none of the EHI runs attains $\mathcal I_{0.1}$ when $d=8$ (10 iterations), and 3 EHI runs out of 10 do not attain $\mathcal I_{0.1}$ when $d=22$ (25 iterations).
	
	In terms of Hypervolume Indicator in $\mathcal I_{0.1}$ (Table \ref{tab:comparaison_qmei_mei}), mEI behaves slightly better than EHI even though the aim of the former is a local attainment of the Pareto front, and not a maximal hypervolume in $\mathcal I_{0.1}$. 2-mEI's performance is not better than mEI at half the budget (which has same wall-clock time but twice less function evaluations). It however exhibits better results than EHI at the same number of iterations (EHI, half budget), and EHI at twice the temporal cost in the case $d=8$. EHI outperforms 2-mEI when $d=22$, but at the cost of 25 more iterations. 
	This may be explained by a weaker estimation of the Pareto front center: we have noticed that with a broader central goal, $\mathcal I_{0.3}$, the Hypervolume indicator is slightly improved for 2-mEI in comparison with mEI, half budget (same wall-clock time): the mechanism of adding $q$ points simultaneously spreads the points a bit more in the targeted region than mEI does. In this broader central part, the time to target also highlights q-mEI over mEI.
	
	In this experiment, the practical performance of mEI and q-mEI is negatively biased 
	in that, because these algorithms estimate the center position, they may drift and miss it. 
	While such convergences would be of practical interest as they yield Pareto optimal solutions, certainly somewhat off-centered, they are counted as negative results.
	This bias will not be present in the following Section, where the targeted area is defined through a user-defined $\RR$.
	
	Last, notice that the standard deviation (in brackets) is slightly lower with the q-mEI than with the mEI, meaning that a more stable convergence to the center occurs. The standard deviations are quite large for these indicators because the part of the Pareto front that is considered ($\mathcal I_{0.1}$) is very narrow; they are smaller in $\mathcal I_{0.3}$ for instance.
	
	\subsubsection{Batch targeting of a user-defined region}
	\label{section:region_targeting}
\begin{figure}[h!]
		\centering
		\includegraphics[width=0.48\textwidth]{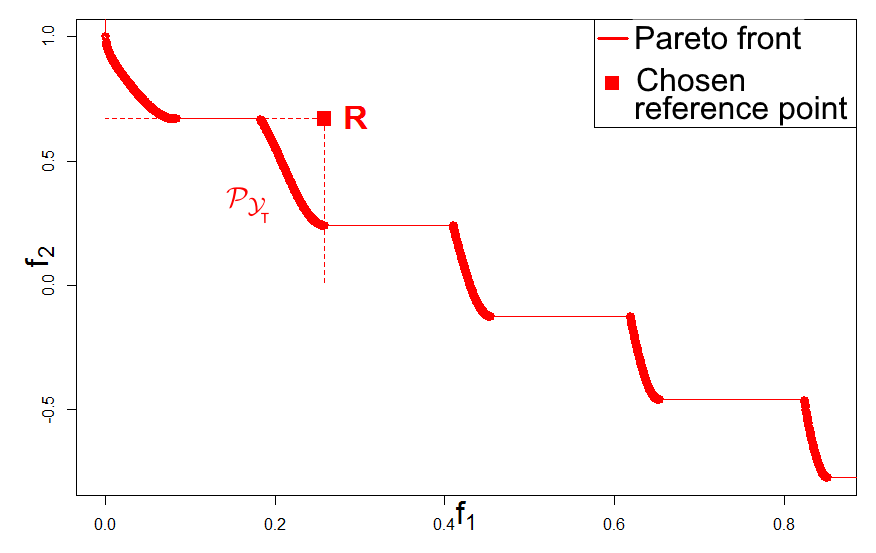}	
		\includegraphics[width=0.51\textwidth]{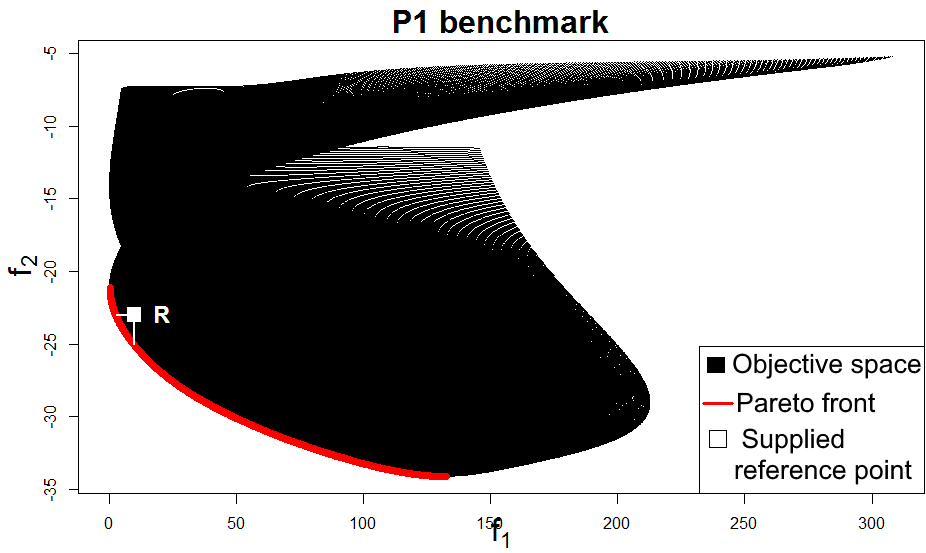}
		\caption{Left: Pareto front of the ZDT3 problem and chosen $\RR$. $\mathcal{P}_{\mathcal Y_T}$ is targeted. Right: objective space and Pareto front of the P1 problem, and targeted region defined through $\RR$.}
		\label{fig:front_zdt3_p1}
	\end{figure}

	Last, let us analyze the ability of q-mEI to attain a region of the Pareto front defined through a reference point $\RR$. Two popular analytical test functions for MOO are considered.
	The first is the ZDT3 \cite{zdt2000a} function which is represented in Figure \ref{fig:front_zdt3_p1}.
	The Pareto set and front of this bi-objective problem consist of five disconnected parts, and we target the second sub-front by setting $\RR$ to its Nadir, $\RR=(0.258,0.670)^\top$.
	In the $d=4$ dimensional version of ZDT3 that we consider in the following experiments, less than 0.003\% of the input space $X=[0,1]^d$ overshoots this target.
	
	In the second experiment, we consider the P1 benchmark problem of \cite{Parr} which is also plotted in Figure \ref{fig:front_zdt3_p1}. It has $d=2$ dimensions, and we target the part of the objective space such that $f_1(\x)\le10$ and $f_2(\x)\le-23$ by setting $\RR=(10,-23)^\top$. This corresponds to approximately 0.9\% of the design space, $X=[0,1]^2$.

	The sequential mEI is compared to q-mEI for batches of $q=2$ and $q=4$ iterates and to two state-of-the-art algorithms: EHI, the Bayesian method already discussed in Section \ref{sec:sequential}, and the evolutionary multi-objective optimization algorithm NSGA-II \cite{NSGAII}. 
	
	In the ZDT3 problem, the Bayesian algorithms (q-mEI, mEI and EHI) start with an initial design of experiments of size $n=20$ and are run for 20 additional iterations. 
	Remember that during these 20 iterations, q-mEI enables to evaluate $\mathbf f(\cdot)$ $20\times q$ times, against 20 function evaluations for mEI and EHI. 
	For NSGA-II, a population of 20 individuals is used, and the results are shown after the second generation (i.e., after 20 additional function evaluations, the budget of mEI and EHI), and after 20 supplementary generations. 
	This second comparison would only be fair if 20 computing nodes were available for evaluating the $\mathbf f(\cdot)$'s in parallel, which is usually not the case.
	However, it will show the advantage of Bayesian methods in terms of function evaluations (20 against 400), already pointed out in \cite{SUR}. 
	Runs of NSGA-II with different number of generations and population sizes have also been investigated but, since they do not change our conclusions and for the sake of brevity, they are not shown here.
	
	For the P1 problem, the size of the initial design of experiments is $n=8$, and the Bayesian algorithms are run for 12 iterations (i.e., 12 function evaluations for EHI and mEI and 24 or 48 evaluations for $q=2$ or 4, respectively). 
	NSGA-II is run for 12 generations with a population of 12 individuals, to enable runtime comparisons.
	
	In order to compare the algorithms at a fixed budget, we do not stop the optimization once the local convergence criterion to the Pareto front is triggered, even though this situation frequently occurs in the mEI and q-mEI runs.
	
	Seven metrics are used for comparison. The first one is the time to target, already employed in Section \ref{section:center_targeting}, which is the number of function evaluations required by an algorithm to dominate $\RR$.
	The second metric is the Hypervolume indicator \cite{TheseZitzler} computed up to $\RR$: it restricts the indicator to the part of the Pareto front which dominates $\RR$.
	The third metric is the number of obtained solutions that dominated $\RR$, i.e., that are in the preferred region. 
	The Euclidean distance to the second Pareto set $\mathcal{P}_{\mathcal X_T}$ (calculated at $\x$ as $\underset{\x' \in \mathcal{P}_{\mathcal X_T}}{\min}\Vert \x - \x'\Vert$ where $\mathbf f(\x') \in \mathcal{P}_{\mathcal Y_T}$)
	is the fourth metric, investigated to address the Pareto-optimality in the targeted region of the obtained designs.
	Last, as the motivation for targeting special parts of the Pareto is to converge more accurately towards some parts of the Pareto front, the Euclidean distance to the whole Pareto set $\PX$ is also compared. This metric allows to discuss distance to Pareto optimality. The pendants of the latter two indicators in the objective space are the Euclidean distance to the targeted Pareto front $\mathcal{P}_{\mathcal Y_T}$, and the distance to the whole Pareto front, $\mathcal{P}_{\mathcal Y}$, respectively.
	These metrics are averaged over 10 runs starting from different initializations, and their standard deviations shown in brackets.
	
	The 8 compared alternatives are mEI, EHI, 2mEI, 2mEI$_t$, 4mEI, 4mEI$_t$,NSGA-II$_1$ and NSGA-II$_{20}$ (or NSGA-II$_{12}$ for the P1 problem). 
	2mEI and 4mEI correspond to the batch version of mEI using $q=2$ or $q=4$ simultaneous iterates and the same number of function evaluations as mEI (20 for ZDT3 or 12 for P1), hence for $q$ times less iterations. 
	The subscript $t$ indicates that the criterion is compared for the same wall-clock time, i.e., the same number of iterations.
	NSGA-II$_1$ corresponds to two generations ($12\times1=12$ or $20\times1=20$ function evaluations) and NSGA-II$_{12}$ and NSGA-II$_{20}$ to fronts obtained after 12 or 20 generations respectively ($12\times12=144$ or $20\times20=400$ function evaluations).
	
	\begin{table}[!ht]
		\centering
		\caption{Comparison of the different infill criteria and algorithms for the ZDT3 function, with respect to the five metrics. The results are averaged over 10 runs, and the standard deviation is shown in brackets. The number of function evaluations for each technique can be found in the row \#$\mathbf f(\cdot)$.
		}
		\setlength\tabcolsep{1pt}
		\makebox[\textwidth][c]{
			\begin{tabu}{|c|c|c|c|c|c|c|c|c|}
				\hline
				Criterion & \multicolumn{2}{c|}{Sequential targeting} & \multicolumn{2}{c|}{Batch targeting, $q=2$} & \multicolumn{2}{c|}{Batch targeting, $q=4$} & \multicolumn{2}{c|}{NSGA-II}\\\hline
				& mEI & EHI & 2mEI & 2mEI$_t$ & 4mEI & 4mEI$_t$ & NSGA-II$_1$ & NSGA-II$_{20}$\\
				\#$\mathbf f(\cdot)$ & 20 & 20 & 2$\times$10 & 2$\times$20 & 4$\times$5 & 4$\times$20 & 20 & 20$\times$20\\\hline
				
				Time to target & 4.2 \scriptsize(2.6) & $\times$ {\color{red}16.7} & 6.3 \scriptsize(4.3) & 6.3 \scriptsize(4.3) & $\times$ {\color{blue}12.5 \scriptsize(6.6)} & 12.5 \scriptsize(6.6) & $\times$ {\color{blue}314.5 \scriptsize(141.3)} & $\times$ {\color{blue}314.5 \scriptsize(141.3)} \\\hline
				
				Hypervolume & 0.634 \scriptsize(0.078) & 0.218 \scriptsize(0.353) & 0.548 \scriptsize(0.201) & 0.621 \scriptsize(0.147) & 0.424 \scriptsize(0.227) & 0.622 \scriptsize(0.088) & 0 & 0.248 \scriptsize(0.253)\\
				
				Distance to $\mathcal{P}_{\mathcal X_T}$ & 0 & 0.097 \scriptsize(0.090) & 0.004 \scriptsize(0.013) & 0 & 0.007 \scriptsize(0.011) & 0.003 \scriptsize(0.010) & 0.369 \scriptsize(0.092) & 0.049 \scriptsize(0.036)\\
				
				Distance to $\PX$ & 0 & 0 & 0.004 \scriptsize(0.013) & 0 & 0.006 \scriptsize(0.011) & 0.003 \scriptsize(0.010) & 0.300 \scriptsize(0.131) & 0.012 \scriptsize(0.010)\\
				
				Distance to $\mathcal{P}_{\mathcal Y_T}$ & 0 & 0.202 \scriptsize(0.190) & 0.002 \scriptsize(0.005) & 0 & 0.006 \scriptsize(0.011) & 0.001 \scriptsize(0.003) & 0.770 \scriptsize(0.437) & 0.045 \scriptsize(0.045)\\
				
				Distance to $\mathcal{P}_{\mathcal Y}$ & 0 & 0 & 0.002 \scriptsize(0.005) & 0 & 0.005 \scriptsize(0.011) & 0.001 \scriptsize(0.003) & 0.638 \scriptsize(0.387) & 0.005 \scriptsize(0.002)\\
				
				Solutions $\prec\RR$ & 4.1 \scriptsize(1.8) & 1.1 \scriptsize(1.9) & 2.8 \scriptsize(1.0) & 3.6 \scriptsize(0.8) & 1.5 \scriptsize(1.0) & 2.4 \scriptsize(1.0) & 0 & 4.2 \scriptsize(4.1)\\\hline
			\end{tabu}
		}
		\label{tab:results}
	\end{table}
	
	\begin{table}[!ht]
		\centering
		\caption{Comparison of the different infill criteria and algorithms for the P1 function with respect to the five chosen metrics. The results are averaged over 10 runs, and the standard deviation is shown in brackets. The number of function evaluations for each technique can be found in the row \#$\mathbf f(\cdot)$.
		}
		\setlength\tabcolsep{1pt}
		\makebox[\textwidth][c]{
			\begin{tabu}{|c|c|c|c|c|c|c|c|c|}
				\hline
				Criterion & \multicolumn{2}{c|}{Sequential targeting} & \multicolumn{2}{c|}{Batch targeting, $q=2$} & \multicolumn{2}{c|}{Batch targeting, $q=4$} & \multicolumn{2}{c|}{NSGA-II}\\\hline
				& mEI & EHI & 2mEI & 2mEI$_t$ & 4mEI & 4mEI$_t$ & NSGA-II$_1$ & NSGA-II$_{12}$\\
				\#$\mathbf f(\cdot)$ & 12 & 12 & 2$\times$6 & 2$\times$12 & 4$\times$3 & 4$\times$12 & 12 & 12$\times$12\\\hline
				
				Time to target & 4.6 \scriptsize(3.5) & $\times$ {\color{red}9.6} & $\times$ {\color{blue}6.2 \scriptsize(4.4)} & 6.2 \scriptsize(4.4) & $\times$ {\color{blue} 8 \scriptsize(6.4)} & 8 \scriptsize(6.4) & $\times$ {\color{blue}73.1 \scriptsize(72.3)} & $\times$ {\color{blue}73.1 \scriptsize(72.3)}\\\hline
				
				Hypervolume & 0.620 \scriptsize(0.165) & 0.163 \scriptsize(0.213) & 0.541 \scriptsize(0.199) & 0.696 \scriptsize(0.051) & 0.459 \scriptsize(0.244) & 0.685 \scriptsize(0.085) & 0.043 \scriptsize(0.136) & 0.394 \scriptsize(0.295)\\
				
				Distance to $\mathcal{P}_{{\mathcal X}_T}$ & 0 & 0.023 \scriptsize(0.020) & 0.009 \scriptsize(0.020) & 0 & 0.025 \scriptsize(0.042) & 0 & 0.107 \scriptsize(0.079) & 0.017 \scriptsize(0.011)\\
				
				Distance to $\PX$ & 0 & 0.002 \scriptsize(0.004) & 0.002 \scriptsize(0.004) & 0 & 0.005 \scriptsize(0.006) & 0 & 0.045 \scriptsize(0.050) & 0.001 \scriptsize(0.002)\\
				
				Distance to $\mathcal{P}_{{\mathcal Y}_T}$ & 0 & 0.98 \scriptsize(1.59) & 0.10 \scriptsize(0.27) & 0 & 0.71 \scriptsize(1.62) & 0 & 4.74 \scriptsize(3.52) & 0.25 \scriptsize(0.34)\\
				
				Distance to $\PY$ & 0 & 0.03 \scriptsize(0.04) & 0.02 \scriptsize(0.03) & 0 & 0.03 \scriptsize(0.04) & 0 & 0.89 \scriptsize(1.29) & 0.01 \scriptsize(0.01)\\
				
				Solutions $\prec\RR$ & 6.5 \scriptsize(2.5) & 0.6 \scriptsize(0.7) & 5 \scriptsize(2.2) & 13.4 \scriptsize(2.3) & 3.3 \scriptsize(1.6) & 13.6 \scriptsize(1.2) & 0.2 \scriptsize(0.6) & 2.8 \scriptsize(2.4)\\\hline
			\end{tabu}
		}
		\label{tab:results_P1}
	\end{table}
	
	These results confirm that mEI is able to attain a user-defined part of the Pareto front within a limited number of iterations. 
	The time to attain the target is much smaller with the targeting criteria than with the others. 
	In the row ``Time to target'', a '$\times$' indicates that at least one run was not able to dominate $\RR$.
	To be more precise, in the ZDT3 problem (Table \ref{tab:results}), 7 EHI runs, 1 4mEI run (after 5 iterations), all 10 NSGA-II$_1$ runs and 3 NSGA-II$_{20}$ runs have not reached $\RR$. 
	In the P1 problem (Table \ref{tab:results_P1}), 5 EHI runs, 1 2mEI run (after 6 iterations), 1 4mEI run (after 3 iterations), 9 NSGA-II$_1$ runs and 2 NSGA-II$_{12}$ runs were not able to reach $\RR$.
	As q-mEI and NSGA-II have been run for more iterations than the budget prescribed in other experiments, the blue color indicates the true value of the indicator if these algorithms were run for more iterations than authorized, until attaining $\RR$. 
	For EHI, the red color indicates the Expected Runtime\footnote{A rough estimator for the Expected Runtime is $\overline{T}_s/p_s$ where $\overline{T}_s$ and $p_s$ correspond to the runtime of successful runs and to the proportion of successful runs respectively 
	} \cite{auger2005performance} to reach $\RR$. 
	In both experiments mEI takes the least function evaluations to attain $\RR$. However, as q-mEI enables to evaluate $q$ designs per iteration, 2mEI and 4mEI require fewer iterations (hence less wall-clock time) than mEI to dominate $\RR$.
	At small number of iterations (e.g. 3 iterations for the 4mEI on the P1 benchmark, corresponding to 12 function evaluations), one 4mEI run out of 10 fails to reach $\RR$, even though mEI with the same number of iterations always attains it. 
	The other indicators confirm that q-mEI is worse than mEI at fixed number of function evaluations, but should be preferred at the same wall-clock time. 
	Indeed, at the same number of function evaluations, q-mEI makes $q$ times fewer metamodel updates to direct the search towards good parts of the design space.
	
	In comparison with EHI, $\mathcal I_\RR$ is attained more consistently by q-mEI and mEI on both test functions, as confirmed by the larger hypervolume and the larger number of solutions that dominate $\RR$.
	The standard deviations of these indicators is also smaller, which means that the results of q-mEI are more repetitive.
	Indeed some EHI runs converge to $\mathcal{P}_{\mathcal Y_T}$ while other runs do not. 
	The distance to $\mathcal{P}_{\mathcal X_T}$ is also reduced (in fact, points belonging to $\mathcal{P}_{\mathcal X_T}$ are found during each mEI and q-mEI run). 
	However, the distance to $\PX$ does not show any benefit of q-mEI and mEI here, since each run of EHI is also able to reach the Pareto front in some other part of it. The same conclusions are made when considering the distance to $\mathcal{P}_{\mathcal Y_T}$ and the distance to $\PY$. Note that for mEI and q-mEI, the distance to $\PY$ equals the distance to $\mathcal{P}_{\mathcal Y_T}$: the Pareto front has been attained the best in the targeted part.
	
	These remarks also hold when comparing mEI and q-mEI with NSGA-II: 
	$\mathcal I_\RR$ is attained more consistently by mEI, as proved by the Hypervolume indicator and the distance to $\mathcal{P}_{\mathcal X_T}$. 
	The distance to $\mathcal{P}_{\mathcal X}$ is also smaller: mEI is able to produce Pareto optimal outcomes, which is not the case for NSGA-II even with a much higher number of function evaluations (400 against 20). 
	More solutions dominating $\RR$ are produced by NSGA-II, which is explained by the much larger number of function evaluations.
	
	\section{Conclusions and future work}
	In this paper, we have described an efficient infill criterion called mEI to guide a multi-objective Bayesian algorithm towards a non-dominated target. 
	We have also singled out one such target by introducing the concept of Pareto front center.
	Numerical experiments have shown that targeting the center of the Pareto front is feasible within restricted budgets for which the approximation of the entire front would not be feasible.
	A multi-point extension to the mEI criterion has been proposed which opens the way to targeted Bayesian multi-objective optimization carried out in a parallel computing environment. 
	At the same temporal cost, $q$ times more experiments can be performed to attain the Pareto front. This criterion, called q-mEI, has been optimized for $q=2$ and 4 batch sizes and is still computed using Monte-Carlo simulations. Experiments with an engineering test bed and standard test functions have confirmed that wall-clock time benefits are achievable with the multi-point infill criterion q-mEI.
	
	mEI aims at rapidly reaching one point of the Pareto front and a test to detect this convergence has been described. A perspective to this work is to use an eventual remaining evaluation budget to broaden the search of the Pareto set.
	Besides, in the spirit of \cite{qEIFormule,marmin2015differentiating,marmin2016efficient}, it might be possible to derive an analytical expression for q-mEI and its gradient, an important perspective as it would allow to optimize it efficiently.

	\section*{Acknowledgments}
	This research was performed within the framework of a CIFRE grant (convention \#2016/0690) established between the ANRT and the Groupe PSA for the doctoral work of David Gaudrie.
	
	\bibliographystyle{plain}
	\bibliography{biblio}

\begin{thebibliography}{10}

\bibitem{auger2009theory}
Anne Auger, Johannes Bader, Dimo Brockhoff, and Eckart Zitzler.
\newblock Theory of the hypervolume indicator: optimal $\mu$-distributions and
  the choice of the reference point.
\newblock In {\em Proceedings of the tenth ACM SIGEVO workshop on Foundations
  of genetic algorithms}, pages 87--102. ACM, 2009.

\bibitem{auger2012hypervolume}
Anne Auger, Johannes Bader, Dimo Brockhoff, and Eckart Zitzler.
\newblock Hypervolume-based multiobjective optimization: Theoretical
  foundations and practical implications.
\newblock {\em Theoretical Computer Science}, 425:75--103, 2012.

\bibitem{auger2005performance}
Anne Auger and Nikolaus Hansen.
\newblock Performance evaluation of an advanced local search evolutionary
  algorithm.
\newblock In {\em 2005 IEEE congress on evolutionary computation}, volume~2,
  pages 1777--1784. IEEE, 2005.

\bibitem{pref2}
Slim Bechikh, Marouane Kessentini, Lamjed~Ben Said, and Khaled Gh{\'e}dira.
\newblock Chap. 4: Preference incorporation in evolutionary multiobjective
  optimization: A survey of the state-of-the-art.
\newblock {\em Advances in Computers}, 98:141--207, 2015.

\bibitem{beume2007sms}
Nicola Beume, Boris Naujoks, and Michael Emmerich.
\newblock Sms-emoa: Multiobjective selection based on dominated hypervolume.
\newblock {\em European Journal of Operational Research}, 181(3):1653--1669,
  2007.

\bibitem{binois2015quantifying}
Micka{\"e}l Binois, David Ginsbourger, and Olivier Roustant.
\newblock Quantifying uncertainty on {P}areto fronts with {G}aussian process
  conditional simulations.
\newblock {\em European Journal of Operational Research}, 243(2):386--394,
  2015.

\bibitem{branke2004finding}
J{\"u}rgen Branke, Kalyanmoy Deb, Henning Dierolf, and Matthias Osswald.
\newblock Finding knees in multi-objective optimization.
\newblock In {\em International conference on parallel problem solving from
  nature}, pages 722--731. Springer, 2004.

\bibitem{branke2008multiobjective}
J{\"u}rgen Branke, Kalyanmoy Deb, Kaisa Miettinen, and Roman Slowi{\'n}ski.
\newblock {\em Multiobjective optimization: Interactive and evolutionary
  approaches}, volume 5252.
\newblock Springer Science \& Business Media, 2008.

\bibitem{buchanan03}
John Buchanan and Lorraine Gardiner.
\newblock A comparison of two reference point methods in multiple objective
  mathematical programming.
\newblock {\em European Journal of Operational Research}, 149(1):17--34, 2003.

\bibitem{qEIFormule}
Cl{\'e}ment Chevalier and David Ginsbourger.
\newblock Fast computation of the multi-points expected improvement with
  applications in batch selection.
\newblock In {\em International Conference on Learning and Intelligent
  Optimization}, pages 59--69. Springer, 2013.

\bibitem{couckuyt2014fast}
Ivo Couckuyt, Dirk Deschrijver, and Tom Dhaene.
\newblock Fast calculation of multiobjective probability of improvement and
  expected improvement criteria for {P}areto optimization.
\newblock {\em Journal of Global Optimization}, 60(3):575--594, 2014.

\bibitem{NSGAII}
Kalyanmoy Deb, Amrit Pratap, Sameer Agarwal, and Tamt Meyarivan.
\newblock A fast and elitist multiobjective genetic algorithm: {NSGA-II}.
\newblock {\em IEEE transactions on evolutionary computation}, 6(2):182--197,
  2002.

\bibitem{deb2006reference}
Kalyanmoy Deb and J~Sundar.
\newblock Reference point based multi-objective optimization using evolutionary
  algorithms.
\newblock In {\em Proceedings of the 8th annual conference on Genetic and
  evolutionary computation}, pages 635--642. ACM, 2006.

\bibitem{EHI}
Michael Emmerich, Andr{\'e} Deutz, and Jan~Willem Klinkenberg.
\newblock Hypervolume-based expected improvement: Monotonicity properties and
  exact computation.
\newblock In {\em IEEE Congress on Evolutionary Computation (CEC), 2011}, pages
  2147--2154. IEEE, 2011.

\bibitem{emmerich2016multicriteria}
Michael Emmerich, Kaifeng Yang, Andr{\'e} Deutz, Hao Wang, and Carlos~M
  Fonseca.
\newblock A multicriteria generalization of {B}ayesian global optimization.
\newblock In {\em Advances in Stochastic and Deterministic Global
  Optimization}, pages 229--242. Springer, 2016.

\bibitem{TheseFeliot}
Paul Feliot.
\newblock {\em Une approche {Bayesienne} pour l'optimisation multi-objectif
  sous contraintes}.
\newblock PhD thesis, Universite Paris-Saclay, 2017.

\bibitem{feliotEWHI2019}
Paul Feliot, Julien Bect, and Emmanuel Vazquez.
\newblock User preferences in {B}ayesian multi-objective optimization: The
  expected weighted hypervolume improvement criterion.
\newblock In Giuseppe Nicosia, Panos Pardalos, Giovanni Giuffrida, Renato
  Umeton, and Vincenzo Sciacca, editors, {\em Machine Learning, Optimization,
  and Data Science}, pages 533--544, Cham, 2019. Springer International
  Publishing.

\bibitem{frazier2012parallel}
Peter~I Frazier and Scott~C Clark.
\newblock Parallel global optimization using an improved multi-points expected
  improvement criterion.
\newblock In {\em INFORMS Optimization Society Conference, Miami FL},
  volume~26, 2012.

\bibitem{book_mcdm}
Tomas Gal, Theodor Stewart, and Thomas Hanne.
\newblock {\em Multicriteria decision making: advances in {MCDM} models,
  algorithms, theory, and applications}, volume~21.
\newblock Springer Science \& Business Media, 1999.

\bibitem{gaudrie2018budgeted}
David Gaudrie, Rodolphe Le~Riche, Victor Picheny, Benoit Enaux, and Vincent
  Herbert.
\newblock Budgeted multi-objective optimization with a focus on the central
  part of the pareto front-extended version.
\newblock {\em arXiv preprint arXiv:1809.10482}, 2018.

\bibitem{Asynchrone1}
David Ginsbourger, Janis Janusevskis, and Rodolphe Le~Riche.
\newblock Dealing with asynchronicity in parallel gaussian process based global
  optimization.
\newblock In {\em 4th International Conference of the ERCIM WG on computing and
  statistics (ERCIM'11)}, 2011.

\bibitem{ginsbourger2010towards}
David Ginsbourger and Rodolphe Le~Riche.
\newblock Towards gaussian process-based optimization with finite time horizon.
\newblock In {\em mODa 9--Advances in Model-Oriented Design and Analysis},
  pages 89--96. Springer, 2010.

\bibitem{qEI}
David Ginsbourger, Rodolphe Le~Riche, and Laurent Carraro.
\newblock Kriging is well-suited to parallelize optimization.
\newblock In {\em Computational Intelligence in Expensive Optimization
  Problems}, pages 131--162. Springer, 2010.

\bibitem{horn2015model}
Daniel Horn, Tobias Wagner, Dirk Biermann, Claus Weihs, and Bernd Bischl.
\newblock Model-based multi-objective optimization: taxonomy, multi-point
  proposal, toolbox and benchmark.
\newblock In {\em International Conference on Evolutionary Multi-Criterion
  Optimization}, pages 64--78. Springer, 2015.

\bibitem{REMOA}
Hisao Ishibuchi, Yasuhiro Hitotsuyanagi, Noritaka Tsukamoto, and Yusuke Nojima.
\newblock Many-objective test problems to visually examine the behavior of
  multiobjective evolution in a decision space.
\newblock In {\em International Conference on Parallel Problem Solving from
  Nature}, pages 91--100. Springer, 2010.

\bibitem{Asynchrone2}
Janis Janusevskis, Rodolphe Le~Riche, and David Ginsbourger.
\newblock {Parallel expected improvements for global optimization: summary,
  bounds and speed-up}.
\newblock Technical report, Institut Fayol, \'Ecole des Mines de
  Saint-\'Etienne, 2011.

\bibitem{janusevskis2012expected}
Janis Janusevskis, Rodolphe Le~Riche, David Ginsbourger, and Ramunas
  Girdziusas.
\newblock Expected improvements for the asynchronous parallel global
  optimization of expensive functions: Potentials and challenges.
\newblock In {\em Learning and Intelligent Optimization}, pages 413--418.
  Springer, 2012.

\bibitem{EGO}
Donald~R Jones, Matthias Schonlau, and William Welch.
\newblock Efficient {G}lobal {O}ptimization of expensive black-box functions.
\newblock {\em Journal of Global optimization}, 13(4):455--492, 1998.

\bibitem{KS}
Ehud Kalai and Meir Smorodinsky.
\newblock Other solutions to {Nash}'s bargaining problem.
\newblock {\em Econometrica: Journal of the Econometric Society}, pages
  513--518, 1975.

\bibitem{Keane}
Andy~J Keane.
\newblock Statistical improvement criteria for use in multiobjective design
  optimization.
\newblock {\em AIAA journal}, 44(4):879--891, 2006.

\bibitem{Parego}
Joshua Knowles.
\newblock {ParEGO}: A hybrid algorithm with on-line landscape approximation for
  expensive multiobjective optimization problems.
\newblock {\em IEEE Transactions on Evolutionary Computation}, 10(1):50--66,
  2006.

\bibitem{marmin2015differentiating}
S{\'e}bastien Marmin, Cl{\'e}ment Chevalier, and David Ginsbourger.
\newblock Differentiating the multipoint expected improvement for optimal batch
  design.
\newblock In {\em International Workshop on Machine Learning, Optimization and
  Big Data}, pages 37--48. Springer, 2015.

\bibitem{marmin2016efficient}
S{\'e}bastien Marmin, Cl{\'e}ment Chevalier, and David Ginsbourger.
\newblock Efficient batch-sequential bayesian optimization with moments of
  truncated gaussian vectors.
\newblock {\em arXiv preprint arXiv:1609.02700}, 2016.

\bibitem{miettinen1998nonlinear}
Kaisa Miettinen.
\newblock {\em Nonlinear multiobjective optimization}, volume~12.
\newblock Springer Science \& Business Media, 1998.

\bibitem{molchanov2005theory}
Ilya Molchanov.
\newblock {\em Theory of random sets}, volume~19.
\newblock Springer, 2005.

\bibitem{pardalos2017non}
Panos~M Pardalos, Antanas {\v{Z}}ilinskas, and Julius {\v{Z}}ilinskas.
\newblock {\em Non-convex multi-objective optimization}.
\newblock Springer, 2017.

\bibitem{Parr}
James Parr.
\newblock {\em Improvement criteria for constraint handling and multiobjective
  optimization}.
\newblock PhD thesis, University of Southampton, 2013.

\bibitem{SUR}
Victor Picheny.
\newblock Multiobjective optimization using {Gaussian} process emulators via
  stepwise uncertainty reduction.
\newblock {\em Statistics and Computing}, 25(6):1265--1280, 2015.

\bibitem{SMS}
Wolfgang Ponweiser, Tobias Wagner, Dirk Biermann, and Markus Vincze.
\newblock Multiobjective optimization on a limited budget of evaluations using
  model-assisted {S}-metric selection.
\newblock In {\em International Conf. on Parallel Problem Solving from Nature},
  pages 784--794. Springer, 2008.

\bibitem{ribaud_phd}
Melina Ribaud.
\newblock {\em Krigeage pour la conception de turbomachines: grande dimension
  et optimisation robuste}.
\newblock PhD thesis, Universit\'e de Lyon, 2018.

\bibitem{sawaragi1985theory}
Yoshikazu Sawaragi, Hirotaka Nakayama, and Tetsuzo Tanino.
\newblock {\em Theory of multiobjective optimization}, volume 176.
\newblock Elsevier, 1985.

\bibitem{TheseSchonlau}
Matthias Schonlau.
\newblock {\em Computer experiments and global optimization}.
\newblock PhD thesis, University of Waterloo, 1997.

\bibitem{TheseSvensson}
Joshua Svenson.
\newblock {\em Computer experiments: Multiobjective optimization and
  sensitivity analysis}.
\newblock PhD thesis, The Ohio State University, 2011.

\bibitem{EMI}
Joshua Svenson and Thomas~J Santner.
\newblock Multiobjective optimization of expensive black-box functions via
  expected maximin improvement.
\newblock {\em The Ohio State University, Columbus, Ohio}, 32, 2010.

\bibitem{triantaphyllou2000multi}
Evangelos Triantaphyllou.
\newblock Multi-criteria decision making methods.
\newblock In {\em Multi-criteria decision making methods: A comparative study},
  pages 5--21. Springer, 2000.

\bibitem{while2012fast}
Lyndon While, Lucas Bradstreet, and Luigi Barone.
\newblock A fast way of calculating exact hypervolumes.
\newblock {\em IEEE Transactions on Evolutionary Computation}, 16(1):86--95,
  2012.

\bibitem{wierzbicki1980use}
Andrzej Wierzbicki.
\newblock The use of reference objectives in multiobjective optimization.
\newblock In {\em Multiple criteria decision making theory and application},
  pages 468--486. Springer, 1980.

\bibitem{wierzbicki1999reference}
Andrzej Wierzbicki.
\newblock Reference point approaches.
\newblock In {\em Gal, T., Stewart, T., Hanne, T. (eds.) Multicriteria Decision
  Making: Advances in MCDM Models, Algorithms, Theory, and Applications}, pages
  237--275. Springer, 1999.

\bibitem{yang2019multi}
Kaifeng Yang, Michael Emmerich, Andr{\'e} Deutz, and Thomas B{\"a}ck.
\newblock Multi-objective {B}ayesian global optimization using expected
  hypervolume improvement gradient.
\newblock {\em Swarm and evolutionary computation}, 44:945--956, 2019.

\bibitem{yang2017computing}
Kaifeng Yang, Michael Emmerich, Andr{\'e} Deutz, and Carlos~M Fonseca.
\newblock Computing 3-{D} expected hypervolume improvement and related
  integrals in asymptotically optimal time.
\newblock In {\em International Conference on Evolutionary Multi-Criterion
  Optimization}, pages 685--700. Springer, 2017.

\bibitem{TEHI}
Kaifeng Yang, Longmei Li, André Deutz, Thomas B{\"a}ck, and Michael Emmerich.
\newblock Preference-based multiobjective optimization using truncated expected
  hypervolume improvement.
\newblock In {\em Natural Computation, Fuzzy Systems and Knowledge Discovery
  (ICNC-FSKD), 2016 12th International Conference on}, pages 276--281. IEEE,
  2016.

\bibitem{zeleny1976theory}
Milan Zeleny.
\newblock The theory of the displaced ideal.
\newblock In {\em Multiple criteria decision making Kyoto 1975}, pages
  153--206. Springer, 1976.

\bibitem{TheseZitzler}
Eckart Zitzler.
\newblock {\em Evolutionary algorithms for multiobjective optimization: Methods
  and applications}.
\newblock Citeseer, 1999.

\bibitem{zdt2000a}
Eckart Zitzler, Kalyanmoy Deb, and Lothar Thiele.
\newblock {Comparison of Multiobjective Evolutionary Algorithms: Empirical
  Results}.
\newblock {\em Evolutionary Computation}, 8(2):173--195, 2000.

\end{thebibliography}
	
\end{document}